\documentclass[pdflatex,sn-mathphys-num]{sn-jnl}

\usepackage{caption}
\usepackage[caption=false]{subfig}
\usepackage{graphicx}%
\usepackage{multirow}%
\usepackage{amsmath,amssymb,amsfonts}%
\usepackage{amsthm}%
\usepackage{mathrsfs}%
\usepackage[title]{appendix}%
\usepackage{xcolor}%
\usepackage{textcomp}%
\usepackage{manyfoot}%
\usepackage{booktabs}%
\usepackage{algorithm}%
\usepackage{algorithmicx}%
\usepackage{algpseudocode}%
\usepackage{listings}%
\RequirePackage{forest}
\useforestlibrary{edges}
\usepackage{tikz}          
\usetikzlibrary{arrows.meta}  
\usepackage{forest}
\useforestlibrary{edges}

\theoremstyle{thmstyleone}%
\theoremstyle{thmstyletwo}%

\theoremstyle{thmstylethree}%

\begin{document}

\title[Article Title]{Modeling Induced Pleasure through Cognitive Appraisal Prediction via Multimodal Fusion}


\author*[1]{\fnm{Nastaran} \sur{Dab}}\email{nastdab@gmail.com}

\author[2]{\fnm{Raziyeh} \sur{Zall}}\email{zall.raziyeh@gmail.com}

\author[3]{\fnm{Mohammadreza} \sur{Kangavari}}\email{kangavari@iust.ac.ir}

\affil{\orgdiv{Department of Computer Engineering}, \orgname{Iran University of Science and Technology}, \orgaddress{ \city{Tehran}, \country{Iran}}}


\abstract{Multimodal affective computing analyzes user-generated social media content to predict emotional states. However, a critical gap remains in understanding how visual content shapes cognitive interpretations and elicits specific affective experiences such as pleasure. This study introduces a novel computational model to infer video-induced pleasure via cognitive appraisal variables. The proposed model addresses four challenges: (1)  noisy and inconsistent human labels, (2) the semantic gap between "positive emotions" and "pleasure," (3) the scarcity of pleasure-specific datasets, and (4) the limited interpretability of existing black-box fusion methods.
		Our approach integrates data-driven and cognitive theory-driven methods, using cognitive appraisal theory and a fuzzy model within an innovative framework. The model employs transformer-based architectures and attention mechanisms for fine-grained multimodal feature extraction and interpretable fusion to capture both inter- and intra-modal dynamics associated with pleasure. This enables the prediction of underlying appraisal variables, thereby bridging the semantic gap and enhancing model explainability beyond conventional statistical associations. Experimental results validate the efficacy of the proposed method in detecting video-induced pleasure, achieving a peak accuracy of 0.6624 in predicting pleasure levels. These findings highlight promising implications for affective content recommendation, intelligent media creation, and advancing our understanding of how digital media influences human emotions.}

\keywords{Multimodal Affective computing, Pleasure, Cognitive Appraisal Theory, Fuzzy Logic, Transformers, Video Affective Content Analysis.}



\maketitle

\section{Introduction}\label{sec1}

The proliferation of online services like YouTube has transformed video into a primary medium for communication and entertainment, leading to an ever-growing volume of video content. This growth directly influences the affective experiences of viewers, who increasingly rely on video consumption to satisfy emotional requirements and modulate their moods, notably for boredom relief. Consequently, accurately analyzing and labeling videos based on their emotional content has become increasingly vital \cite{hanjalic2005affective, xu2014three, canini2012affective, nath2020comparative, zhu2020affective, yi2024emotion}. The practical utility of such analysis spans from narrative optimization in filmmaking and mood-based content filtering for end-users to the strategic deployment of targeted advertising by distributors. Beyond content creation, understanding the brain mechanisms underlying pleasure and displeasure is a critical goal in emotional neuroscience, as the capacity for normal pleasure is fundamental for healthy psychological functioning \cite{berridge2015pleasure}. Conversely, emotional disorders can manifest as a pathological lack of pleasure or excessive unpleasant feelings like pain, depression, or anxiety \cite{ berridge2011building, moccia2018experience, kringelbach2009towards}.
	
	Despite the clear significance of emotional monitoring, several key challenges remain unresolved in how we capture these states. Recent studies have attempted to infer emotions by analyzing users' spontaneous reactions \cite{nath2020comparative, hassouneh2020development, li2022novel}, often relying on physiological signals such as EEG \cite{huang2021differences, algarni2022deep, liao2024exploring, zhang2024tpro}. However, these approaches are impractical for real-world deployment due to the need for costly, intrusive hardware that not only disrupts natural viewing behavior but also limits scalability and user comfort. To circumvent these physical constraints and move toward more scalable solutions, researchers have shifted focus toward stimulus-based methods using machine learning or deep learning techniques \cite{xu2024infer, wagner2023dawn}.
	
	Stimulus-based methods extract emotional cues straight from the audio-visual content of the video itself \cite{ dudzik2020blast, yi2019affective, xie2024emovit, li2024temporal}. Early efforts in this direction used traditional machine learning with handcrafted features \cite{chan2010affect,irie2010affective}. While functional, these models depend on shallow, linear architectures that struggle to capture the rich, non-linear patterns in large-scale, complex video data, leading to limited accuracy and poor generalization. Moreover, manual feature engineering lacks adaptability: the same hand-designed extractors cannot be easily repurposed across diverse emotion recognition tasks \cite{kratzwald2018deep,younis2024machine, hazmoune2024using}. To overcome the rigidity of manual features, deep neural networks have emerged as the dominant paradigm, capable of learning complex representations directly from raw data\cite{zhang2024deep, latif2021survey, cui2023deep}.
	
	However, the rise of deep learning has introduced a new set of conceptual and structural dilemmas. These models automatically learn hierarchical representations directly from raw audio-visual inputs, with performance driven by the design of the network architecture and end-to-end training, eliminating the bottlenecks of manual intervention and enabling robust, scalable affective analysis. Yet, this performance often comes at the cost of interpretability, as DNNs operate as "black boxes" that ignore the underlying cognitive processes of the viewer \cite{kumar2024interpretable, cortinas2023toward, scherer2018appraisal}. According to Cognitive Appraisal Theory, emotions arise from an individual's subjective evaluation of a situation based on its personal significance and coping potential \cite{lazarus1991emotion, zall2022comparative, zall2025intelligent}. This framework suggests that the emotional response is not triggered directly by the event itself, but by a set of cognitive assessments, such as novelty, goal relevance, agency, and desirability, that determine the event's meaning to the individual \cite{zall2024towards, scherer2001appraisal}.
	
	The disregard for these cognitive foundations in purely data-driven DNNs leads to a critical "semantic gap" in how we define and measure emotional states. Specifically, most existing models rely on large, human-labeled datasets which are often influenced by the subjective and vague nature of human emotion perception. This reliance can lead to inconsistencies and noise in labeling, resulting in the "ground truth" problem where models recognize perceived facial expressions rather than actual internal states \cite{tian2017recognizing, li2024temporal, dudzik2020blast}. While it is established that all human emotions are fundamentally rooted in cognitive appraisal variables, the omission of these factors in current models results in a significant loss of information during the pleasure inference process. Since these appraisal variables are the primary drivers behind the elicitation of any emotional state, ignoring them means neglecting the causal mechanisms that link a video stimulus to a feeling of pleasure. By leveraging Cognitive Appraisal Theory, we can channel and structure this missing information to achieve a more precise and theory-grounded detection of pleasure. 
	Although both discrete emotions and dimensional models are ultimately grounded in cognitive appraisal processes, much of the current research still leans heavily on categorical labels, which can hide the subtler dynamics of valence and arousal. In particular, discrete approaches often treat positive emotions as broad, undifferentiated categories such as "happiness" or "joy," without fully capturing the continuous nature of pleasure \cite{russell2003core, posner2005circumplex}. Pleasure, however, is not a single, dichotomous state; it exists along a graded, continuous dimension of core affect that can shift in intensity and quality depending on different appraisal patterns \cite{russell2003core}. By focusing solely on high-level labels and bypassing frameworks like Valence-Arousal-Dominance (VAD), these models lose the fine-grained detail needed to differentiate varying degrees of pleasure \cite{mehrabian1996pleasure, bakker2014pleasure}. As a result, it becomes harder to tailor video content to evoke specific shades of positive affect. Shifting toward a dimensional view, in which pleasure is explicitly modeled as a measurable, primary axis, helps close the gap between coarse emotional categories and the richer, more nuanced internal experience of the viewer \cite{wrobel2025proxy, jia2025bridging}.
	Moreover, the disregard for cognitive appraisals, combined with the opaque nature of deep learning models, reduces their explainability and limits our understanding of why a particular emotion, like pleasure, is predicted. The lack of datasets explicitly labeled for pleasure, rather than general discrete emotions, further exacerbates these analytical challenges.
	
	This research aims to address these critical gaps by developing an innovative computational model for video-induced pleasure. We leverage advanced approaches, including Transformer-based architectures and multimodal fusion, to reduce reliance on subjective and noisy human labels. By integrating cognitive appraisal theory with a fuzzy model, we propose a novel computational framework that predicts specific appraisal variables (e.g., likelihood, desirability), thereby bridging the semantic gap between "positive emotions" and "pleasure." This approach enhances explainability by providing a causal and interpretable framework for emotion production, moving beyond mere statistical correlations. To overcome data scarcity, we utilize a public dataset with specific appraisal variable labels and further extend it with human-labeled pleasure annotations. Our comprehensive, end-to-end framework optimizes feature extraction and fusion, enhancing accuracy and generalizability while addressing the impracticality of physiological sensors for real-world applications. Experimental results on the EmoStim dataset \cite{somarathna2023emostim} demonstrate the effectiveness of this approach, achieving a peak accuracy of 0.6624 in predicting induced pleasure levels.
	
	The key contributions and innovations of this study on inference video-induced pleasure are as follows:
	\begin{enumerate}
		\item Proposing a novel computational model for video-induced pleasure: This study introduces the first computational model specifically designed to quantify video-induced pleasure, distinguishing it from previous works that primarily identified general emotions. Our model analyzes pleasure, integrating fuzzy appraisal and dimensional theories to offer a new method for personalizing video content based on its pleasure profile.
		\item Establishing an Interpretable, Theory-Driven Framework: By synergizing fuzzy logic with cognitive appraisal and dimensional theories, this study provides a transparent mechanism for emotion reasoning. Unlike opaque deep-learning models, our approach establishes an interpretable bridge between objective video features and subjective pleasure. This integration overcomes the limitations of purely data-driven correlations by offering a causal and theory-anchored explanation for induced emotional responses.
		
		\item Enrichment and Extension of the EmoStim Dataset for Pleasure Inference: To overcome the scarcity of pleasure-specific labels in existing affective resources, we extended the EmoStim dataset \cite{somarathna2023emostim} by conducting a systematic human-labeling phase. By augmenting the original appraisal-rich clips with gold-standard pleasure and displeasure annotations, we curated a comprehensive multimodal resource that facilitates the direct mapping of cognitive appraisals to induced pleasure states.
	\end{enumerate}
	
	To the best of our knowledge, this study presents the first framework to explicitly map Scherer’s CPM appraisal variables onto induced pleasure states within a multimodal context. By leveraging a hierarchical fuzzy inference mechanism to analyze video-driven stimuli, this approach establishes a new baseline for pleasure-centric affective computing and provides a structured methodology for quantifying emotional experiences in rich media.
	
	The remainder of this paper is organized as follows. Section \ref{sec:related_works} reviews related works. Section \ref{sec:methodology} details the framework overview. Section \ref{sec:experiments} describes the experimental setup and results. Finally, Section \ref{sec:conclusion} concludes the paper and outlines future works.

\section{Related Works}\label{sec:related_works}
		
	This section reviews existing literature related to video-induced emotion and pleasure, organized according to their theoretical foundations and methodological approaches. We begin by exploring psychological theories that define pleasure. Next, we analyze computational models for detecting emotion and pleasure from video content. This includes distinguishing between content-centric methods, which focus on audio-visual elements, and user-centric methods, which utilize physiological responses from viewers. We also discuss the advancements toward sophisticated deep neural network-based architectures for feature extraction and integration.
	
	\subsection{Theoretical Perspectives on Emotion and Pleasure}
	
	A wide range of theories have been developed to explore the nature, structure, and effects of pleasure, drawing from philosophical, psychological, and neuroscientific traditions. Each perspective sheds light on different facets of this multifaceted experience, including its emotional, motivational, and cognitive dimensions, as well as its influence on human behavior and well-being. The following subsections review these key theoretical frameworks to provide a clearer and more comprehensive picture of pleasure, its varied manifestations, underlying mechanisms, and broader psychological significance.
	
	\subsubsection{Appraisal Theories} One of the most influential theories in psychology and affective science concerning emotion elicitation is cognitive appraisal theory \cite{zall2024towards}. Appraisal theories conceptualize emotions as the product of cognitive evaluations of stimuli or events \cite{moors2013appraisal}. Individuals assess situations along multiple appraisal dimensions, and the resulting configuration determines the specific emotion experienced. Thanks to their structured and interpretable nature, appraisal-based models can form a cornerstone of computational affective systems, underpinning numerous emotional agents and affective computing frameworks \cite{sullins2024investigating, soleymani2016detecting, barradas2025dynamic}.
	Early theories varied in the number and nature of appraisal variables. Frijda \cite{frijda1989relations} highlighted dimensions such as familiarity, predictability, value, controllability, agency, certainty, and event significance. Scherer \cite{scherer2001appraisal,scherer1993studying} offered a finer-grained account, emphasizing predictability, urgency, power, normative significance, and others, yielding systematic mappings from appraisal patterns to discrete emotions. However, the model's 22 variables often render it impractical for many computational implementations.
	In contrast, the OCC model \cite{ortony2022cognitive} operates with only eight appraisal variables, which has made it one of the most widely adopted frameworks in affective computing. Similarly, Scherer's Component Process Model views emotions as emerging from the continuous, dynamic interaction of subprocesses along dimensions such as relevance, implications, coping potential, and normative significance, ultimately producing a unified emotional experience \cite{somarathna2023emostim}.
	Recent research has extended these foundations to stimulus-induced and media-driven emotions. Appraisal patterns have been successfully applied to model affective responses to music \cite{juslin2025major} and creative-evaluation events \cite{friedrich2025emotional}, demonstrating how external stimuli elicit nuanced pleasure and complex emotions. Building on core mechanisms, other work has systematically incorporated modulating factors, such as personality, mood, and cultural norms, into appraisal models without compromising theoretical integrity \cite{castellanos2026systematic}.
	Further advances have investigated appraisal processes in large language models, uncovering reliable prediction of induced emotions alongside limitations, particularly in control and power dimensions \cite{tak2025aware}. In parallel, deep learning approaches have integrated primary and secondary appraisal components to predict emotions in multi-party conversations \cite{xu2025conversational}. Complementing these, nonlinear dynamical models (e.g., NARX) aligned with Scherer's framework have estimated emotion intensity and temporal dynamics from physiological signals, providing interpretable insights into appraisal-driven changes over time \cite{barradas2025dynamic}.
	Collectively, these developments evolve classical appraisal theories toward more robust, interpretable, and ecologically valid computational accounts, effectively addressing induced emotions in dynamic, real-world, and temporal contexts. These efforts build on earlier frameworks by addressing both stimulus-induced contexts and time-varying emotion changes.
	
	\subsubsection{Dimensional Theories} Dimensional theories \cite{krzeminska2025} represent emotions as continuous points in a low-dimensional space, typically defined by core axes such as valence (pleasure–displeasure) and arousal (activation–deactivation), rather than as discrete categories \cite{ekman1992argument}. The most influential formulation is Russell's circumplex model of affect \cite{russell1980circumplex}, which arranges affective states in a circular pattern within the valence-arousal plane \cite{posner2005circumplex}. Reisenzein \cite{reisenzein1994pleasure} demonstrated that emotional intensity correlates closely with positions along these axes: higher pleasure drives stronger positive states (e.g., happiness), while elevated arousal underpins states like alertness \cite{taverner2021fuzzy}.
	Dimensional representations have proven highly effective for continuous emotion prediction across diverse datasets. Early and mid-range successes include LSTM-based valence-arousal regression on DEAP \cite{liao2024exploring}, LEDPatNet19 on DREAMER \cite{zhang2024tpro}, and Bi-LSTM models on AMIGOS \cite{li2020exploring}, underscoring the practical utility of valence-arousal (and occasionally dominance) spaces in affective computing.
	Recent advances have extended dimensional modeling to multimodal and real-world scenarios. MAVEN employs multimodal attention to predict valence-arousal in conversational videos \cite{ahire2025maven}, while TPC leverages triple-path consistency for emotion distribution modeling from audio \cite{hu2026target}. Complementary efforts include VAD-conditioned speech synthesis \cite{liu2025emotional}, spatiotemporal analysis of bodily expressions for continuous VAD prediction \cite{yu2026emotion}, and deep learning-based valence-arousal estimation from affective images \cite{priyadarshani2024predicting}. These developments collectively advance the capture of nuanced, time-varying, and cross-modal affective states, reinforcing dimensional approaches as a robust and scalable framework for ecologically valid affective computing.
	\subsection{Video-Induced Emotion and Pleasure}
	Research on video-induced emotion and pleasure seeks to infer the affective content of videos using a variety of approaches. This section reviews deep neural network-based models, with a particular focus on methods that directly analyze audio-visual features and those that infer emotions from users' spontaneous reactions.
	\subsubsection{Deep-Neural-Net-Based Models}
	
	While hand-crafted features offer high customizability, their shallow and often linear structures limit effectiveness on large-scale, complex video data, frequently resulting in suboptimal performance. Consequently, deep neural networks have emerged as the dominant paradigm in video-induced emotion recognition, enabling end-to-end hierarchical feature learning directly from raw audio-visual inputs.
	A persistent challenge in this domain is label noise in datasets, which impairs supervised training. Zhu et al. \cite{zhu2020affective} tackled this through MMDQEN, a multimodal deep quality embedding network that infers latent, higher-quality labels from noisy samples, thereby improving emotion classifier accuracy.
	To better exploit multimodal and temporal dynamics, early efforts such as Yi et al.'s AFRN \cite{yi2019affective} incorporated dedicated layers for robust feature extraction, temporal fusion, and cross-modal integration. Building on attention mechanisms, Thao et al. \cite{thao2021attendaffectnet} introduced AttendAffectNet, which leverages self-attention to model inter-modal and temporal dependencies, achieving strong results in predicting viewers' affective responses to movies, particularly in its feature-based variant.
	More recent models have addressed long-range dependencies and data scarcity. Yi et al. \cite{yi2024emotion} proposed LRCANet, which includes a spatiotemporal correlation-aware block to capture relationships across distant input tokens. Similarly, Zhang et al. \cite{zhang2024mart} developed MART, a masked autoencoder-based framework that learns robust representations via masked reconstruction and employs inter-modal attention in a complementary block to mitigate limited training data.
	Other advances target noise suppression and modality-specific enhancements. Li et al. \cite{li2024temporal} presented TE, a mechanism that amplifies relevant temporal information from multimodal states while suppressing irrelevant cues, thereby reducing temporal-domain noise.
	Recent work has also integrated large multimodal models. Guo et al. \cite{guo2025stimuvar} introduced StimuVAR, a spatiotemporal stimuli-aware approach that overcomes the semantic-content bias of multimodal LLMs by incorporating frame-level and token-level awareness for more precise video affective reasoning.
	Foundation-model-oriented efforts further unify perception and reasoning. VidEmo \cite{zhang2025videmo} employs affective-tree reasoning in video foundation models, integrating attribute perception, expression analysis, and high-level emotional understanding via curriculum learning and reinforcement. The MER 2025 challenge \cite{lian2025mer} advances toward LLM-driven generative methods, introducing tracks for semi-supervised, fine-grained, and multimodal emotion recognition to improve interpretability and reliability beyond conventional categorical paradigms.
	
	\subsubsection{Reaction-Based Methods}
	Recent advances in video affective content analysis have shifted toward inferring a video's emotional impact by capturing viewers' spontaneous reactions, particularly through physiological signals such as EEG, rather than relying solely on content features.
	EEG-based emotion recognition has leveraged deep learning within dimensional frameworks to achieve robust prediction of valence, arousal, and dominance. Prominent examples include hybrid CNN-SVM architectures \cite{topic2021emotion}, BiDCNN models \cite{huang2021differences}, and 2D CNN approaches \cite{dudzik2020exploring}, which have demonstrated strong performance on widely used benchmarks including DEAP, DREAMER, SEED, and AMIGOS.
	A key distinction in this domain is between perceived emotions (those conveyed by the media) and induced emotions (those actually felt by the viewer). Tian et al. \cite{tian2017recognizing} highlighted this difference in film-viewing contexts and proposed an LSTM-based fusion model that integrates multimodal information to more accurately capture induced affective states.
	Building on these foundations, Liao et al. \cite{liao2024exploring} explored the relationship between emotional experiences and physiological responses during short social video viewing. They collected multimodal signals, including EEG, galvanic skin response, skin temperature, and heart rate, across seven discrete emotional categories, applied machine learning classifiers to uncover content-physiology links, and publicly released a new dataset to support further research in stimulus-induced affect.
	More recent efforts have targeted subtle emotional dynamics. Zhang et al. \cite{zhang2024tpro} introduced TPRO-NET, an architecture that extracts differential entropy and enhanced features, combining convolutional layers with improved transformer encoders to classify emotions along valence, arousal, and dominance dimensions on DEAP and DREAMER, thereby better capturing nuanced shifts in viewers' responses.
	\subsubsection{Stimulus-Based Methods}
	Despite significant advances in reaction-based approaches, the high cost, intrusiveness, and disruption of natural viewing behavior limit scalability for real-world applications. Consequently, a growing body of work has shifted toward inferring video-induced emotions directly from the audio-visual stimulus content itself.
	Early efforts in this direction focused on closing performance gaps in specific dimensions. Wagner et al. \cite{wagner2023dawn} extensively evaluated Transformer-based audio encoders (wav2vec, HuBERT) and demonstrated substantial improvements in predicting induced pleasure, addressing the longstanding pleasure gap in speech emotion recognition that transfers effectively to video contexts.
	Multimodal fusion has also proven effective for discrete emotion classification. Antonov et al. \cite{antonov2024decoding} developed a convolutional neural network that integrates visual and auditory features to classify eight dominant emotions in viewers' responses to video advertisements, leveraging a large-scale dataset. Other studies have extended stimulus-based inference to include user-generated responses and contextual interplay. Xu et al. \cite{xu2024infer} introduced multimodal sentiment analysis of comment responses to emotionally charged videos, releasing the large-scale CSMV dataset and proposing a multi-view attention mechanism that fuses video and textual cues. Complementing this, Tian et al. \cite{tian2017recognizing} explored the interaction between dialogue and aesthetic elements in films, proposing a multimodal framework to better predict truly induced emotions rather than merely perceived emotions. A parallel and increasingly influential line of research emphasizes personal contextual factors, particularly viewer-specific memories, which conventional content-only models often overlook. Dudzik et al. \cite{dudzik2020blast} showed that textual descriptions of video-triggered personal memories explain substantial inter-individual variability in affective responses; integrating these with audio-visual features markedly improves prediction accuracy. In follow-up work \cite{dudzik2020exploring}, the same authors demonstrated that ambiguous facial expressions become far more predictive when enriched with memory- and video-based context via multimodal machine learning. Building directly on these findings, Dudzik et al. \cite{dudzik2021collecting} introduced the Mementos dataset, a multimodal resource designed for context-sensitive modeling of affect and memory processing in video responses. Unlike prior corpora, Mementos includes the eliciting videos, physiological signals, continuous valence-arousal annotations, detailed personal memory descriptions, and participant demographics/personality traits, enabling more ecologically valid modeling of individual differences. Similarly, Kamran et al. \cite{kamran2023emodnn} proposed EmoDNN, a deep neural network ensemble that jointly infers latent individual factors (e.g., personality, cognition) from short texts while extracting emotions via dynamic dropout convnets, underscoring the value of user-specific cognitive cues even in noisy social media content.
	\subsection{Limitations and Research Gaps}
	
	Although previous works in video emotional content analysis offer valuable contributions, such as personalizing predictions via memory analysis with NLP and Transformers \cite{dudzik2020blast, dudzik2020exploring, dudzik2021collecting}, and improving multimodal emotion detection through spatio-temporal integration in EEG-based models \cite{liao2024exploring, zhang2024tpro, li2020exploring, topic2021emotion, huang2021differences}, several critical limitations persist.
	Many approaches rely heavily on subjective, noisy human labels, leading to the well-known "ground truth" problem where models often capture perceived facial expressions rather than truly induced internal states \cite{tian2017recognizing, li2024temporal}. Moreover, the predominant focus on discrete categorical emotions (e.g., happiness) tends to overlook the semantic gap between broad positive labels and the continuous, graded nature of pleasure, bypassing dimensional frameworks like Valence-Arousal-Dominance (VAD) and losing fine-grained differentiation of pleasurable experiences \cite{russell2003core}. The black-box nature of deep neural networks further compounds this issue, as they frequently ignore the underlying cognitive appraisal processes that causally drive emotional responses, including pleasure elicitation \cite{kumar2024interpretable, cortinas2023toward, scherer2018appraisal}. Similarly, Rahmani et al. \cite{rahmani2023transfer} highlighted that most multimodal approaches overlook user-specific cognitive cues such as personality, proposing a transfer-based adaptive tree that clusters users and transfers knowledge via attention fusion to better account for individual variability. EEG-based methods, while accurate, remain impractical for real-world use due to expensive and intrusive hardware that disrupts natural behavior and limits scalability. These gaps highlight the need for a more theory-grounded, explainable, and scalable model that explicitly incorporates appraisal mechanisms to bridge coarse categories with nuanced affective states.
	
	\section{Methodology}\label{sec:methodology}
	
	We propose CogniPleasure, a novel framework for predicting induced pleasure from multimodal video data, integrating cognitive appraisal theory with advanced data-driven techniques. The CogniPleasure framework consists of two submodules: the Cognitive Appraisal Fusion Module (CAFM) for extracting and fusing multimodal features to predict appraisal variables, and the Fuzzy Pleasure Inference Module (FPIM) to compute a pleasure metric, as illustrated in Figure \ref{fig:1}. The following subsections detail the methodology in three parts: task definition, framework overview, and the specific mechanisms of the CAFM and FPIM submodules, including feature extraction, fusion, and inference processes.
	
	\subsection{Task Definition}\label{sec:task}
	
	Given a collection of videos, each video comprises a set of utterances. Let $U = \{u_1, u_2, \dots, u_N\}$ be a set of $N$ utterances, where each utterance $u_i$ is a multimodal data sample that comprises corresponding audio, visual, and text data streams. CogniPleasure aims to predict the \textbf{Pleasure} value, denoted by $P_i \in \mathbb{R}$, for a specific target utterance $u_i$. To achieve this, our framework infers a set of seven intermediate cognitive appraisal and affective variables, $V_i = \{v_{1_i}, v_{2_i}, \dots, v_{7_i}\}$, from the multimodal input features. These variables provide supplementary cues in predicting induced pleasure. The final evoked pleasure value $P_i$ is obtained according to patterns of these variables by using fuzzy rules.
	
	\subsection{Framework overview}\label{sec:framework}
	\begin{figure*}[t!] 
		\centering 
		\includegraphics[width=0.97\textwidth]{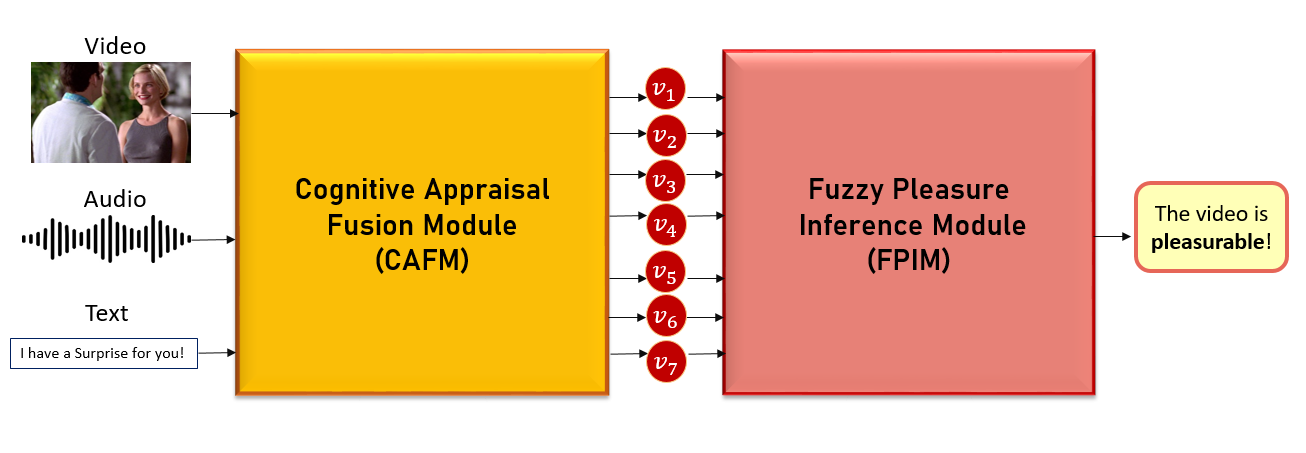} 
		\caption{
			The proposed framework, where V1--V5 denote the predicted appraisal variables from multimodal fusion in CAFM, and V6--V7 represent the calm and boredom emotions used as inputs to the fuzzy model in FPIM for induced pleasure prediction.} 
		\label{fig:1} 
	\end{figure*}
	
	The primary objective of the CogniPleasure framework is to model induced pleasure from multimodal video data by integrating advanced data fusion techniques with robust cognitive appraisal theory, followed by fuzzy logic-based pleasure estimation. The framework aims to address challenges in subjective emotion annotation and enhance the interpretability of pleasure computation by leveraging both data-driven and cognitive theory-driven methodologies.
	Figure \ref{fig:1} provides a comprehensive overview of the proposed framework. The process begins with the Cognitive Appraisal Fusion Module (CAFM), which extracts and fuses meaningful feature representations from raw multimodal data (i.e., audio, text and visual inputs) to predict cognitive appraisal variables such as desirability, controllability, and likelihood. This module serves as the foundation for generating a robust feature set that captures intra-modal and inter-modal interactions, enabling a comprehensive understanding of the emotional context.
	Subsequently, the Fuzzy Pleasure Inference Module (FPIM) takes these predicted appraisal and affective variables as input to compute induced pleasure. The goal of this module is to translate the cognitive appraisals into a numerically and symbolically interpretable pleasure metric using a fuzzy rule-based model, thereby addressing the black-box limitations of previous data-driven approaches. Detailed mechanisms and implementations of each module are elaborated in the subsequent subsections, where the specific processes and optimizations are discussed in depth.
	
	\subsection{The Cognitive Appraisal Fusion Module (CAFM)}
\begin{figure*}[t!] 
		\centering 
		\includegraphics[width=0.999\textwidth]{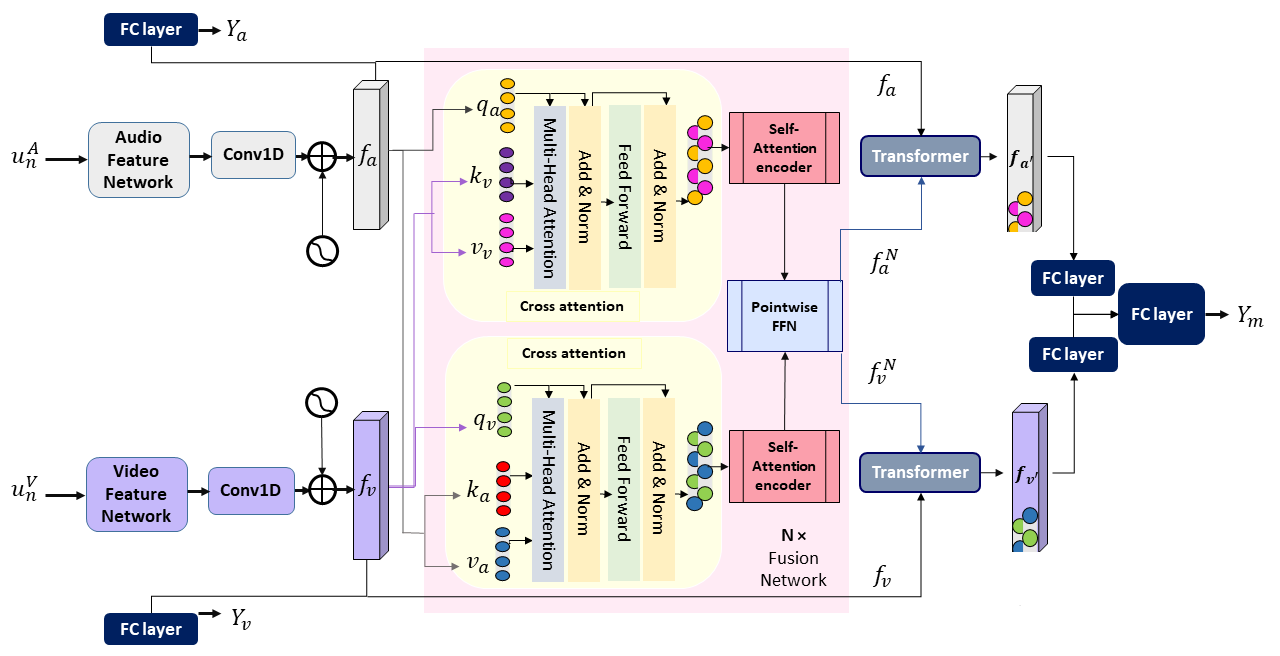} 
		\caption{The Cognitive Appraisal Fusion Module (CAFM) Architecture.} 
		\label{fig:CAFM_overview} 
	\end{figure*}

	The Cognitive Appraisal Fusion Module (CAFM) is designed to robustly predict appraisal variables by processing and integrating multimodal inputs. As illustrated in Figure \ref{fig:CAFM_overview}, the module receives raw text, audio and video streams and outputs the estimated values of the targeted appraisal dimensions. These appraisal variables play a central role in capturing the emotional context embedded within multimodal data and in identifying the cognitive cues that shape the interpretation of perceived situations. Inspired by the cognitive appraisal framework proposed by Sander \cite{sander2005systems}, the predicted variables include:
	\begin{itemize}
		\item \textbf{$V_e$} ({$Expectedness$}): This variable indicates the observer's expectation level regarding the event.
		\item \textbf{$V_l$} ({$Likelihood$}): This variable represents the estimated probability of the event occurring in the utterance $u_i$.
		\item \textbf{$V_d$} ({$Desirability$}): This variable shows how desirable or undesirable the event is for the observer, which can be positive or negative.
		\item \textbf{$V_a$} ({$Agency$}): Indicates who is responsible for the event. In our proposed scenario, this variable is always equivalent to ``other'' because a video viewer does not influence the events occurring in the video.
		\item \textbf{$V_{co}$} ({$Controllability$}): This reflects the number of feasible plans the observer considers in response to the event. In our scenario, it measures the viewer's desire to change the unfolding events in the video.
		\item \textbf{$V_{ca}$} represents the emotion of {$Calm$}, capturing the observer's state of tranquility or relaxation in response to the event.
		\item \textbf{$V_b$} represents the emotion of {$Boredom$}, indicating the observer's lack of engagement or interest in the event.
	\end{itemize}
	These variables form the set $V_i = \{v_{e_i}, v_{l_i}, \dots, v_{b_i}\}$, where $V_i \in \mathbb{R}^7$. They are predicted by integrating multimodal features as detailed below.
	\subsubsection{Representation Learning}
	Our end-to-end architecture overcomes the drawbacks of hand-crafted feature engineering by integrating feature extraction and model training in a unified optimization process, harnessing the power of pre-trained backbones to deliver a robust, cohesive approach for multimodal emotion representation learning.
	We first extract a set of representative features for each modality. Given an utterance $u_i$, let the extracted feature vectors be $F_{\text{audio}_i} \in \mathbb{R}^{d_a}$, $F_{\text{visual}_i} \in \mathbb{R}^{d_v}$, $F_{\text{text}_i} \in \mathbb{R}^{d_t}$ where $d_a$, $d_v$, and $d_t$ are the dimensions of the respective feature vectors. To make our modalities sufficiently perceptible, we pass the obtained feature vectors through a one-dimensional temporal convolution layer.
		
	\begin{equation}
		U_m^* = \text{Conv1D}\left(U_m; k_m\right) \in \mathbb{R}^{T_m \times d},
	\end{equation}
	where $\text{Conv1D}(\cdot)$ denotes the one-dimensional temporal convolution function, $k_m$ represents the kernel size for modality $m$, $U_m$ is the input sequence of modality $m$, $d$ is the unified feature dimension, and $T_m$ indicates the sequence length of modality $m$; here, $m \in \{a, t, v\}$.
	To incorporate temporal ordering into the sequences, we follow the approach of Vaswani et al. \cite{vaswani2017attention} and add sinusoidal position embeddings (PE) to the convolved representations $U_m^*$ as
	\begin{equation}
		\tilde{U}_m = U_m^* + \text{PE}(T_m, d),
	\end{equation}
	where $\text{PE}(T_m, d) \in \mathbb{R}^{T_m \times d}$ computes a unique sinusoidal embedding for each temporal position in the sequence. The function $\text{PE}(\cdot)$ generates fixed positional encodings, with $m \in \{a,t, v\}$. 
	
	Subsequently, we apply cross-modal attention to the position-augmented sequences. This mechanism enables each modality to attend to temporally aligned features from the other modality, thereby injecting complementary contextual cues into its own representations. Such bidirectional information flow fosters synergistic integration, enhancing the model's capacity to capture emotionally coherent patterns that emerge from the interplay between modality dynamics.
	
	\subsubsection{Fusion}
	The fusion module integrates information from multiple modalities. The extracted features are fed into a learning model, represented by a function $\mathcal{M}$, to predict the set of intermediate appraisal and affective variables $V = \{v_{e}, v_{l}, \dots, v_{b}\}$. This can be formally expressed as:
	\begin{equation}
		V_i = \mathcal{M}(F_{\text{audio}}, F_{\text{visual}}, F_{\text{text}})
	\end{equation}
	where $V_i \in \mathbb{R}^7$ and $ i \in \{e, l, \dots, b\} $ indexes the seven intermediate cognitive appraisal and affective variables.
	
	The primary objective of the training phase is to optimize the model $\mathcal{M}$ to minimize the prediction error for the appraisal and affective variables $V_i$ for each utterance in the dataset $U$. The fusion module is divided into three primary components, as depicted in the Figure \ref{fig:CAFM_overview}.
	First, a cross-attention encoder, introduced by \cite{vaswani2017attention}, is employed, analogous to a self-attention encoder. It utilizes a query from one modality and generates keys and values from another. This cross-modal interaction captures inter-modal dependencies, contributing to a more comprehensive understanding of the data. This encoder is formally defined as follows:
	
	\begin{equation} \label{eq:attention_formula}
		Attention(Q_{m1}, K_{m2}, V_{m2}) = \text{softmax}\left(\frac{Q_{m1}K_{m2}^T}{\sqrt{d_k}}\right)V_{m2}
	\end{equation}
	where:
	\begin{align} \label{eq:qkv_definitions}
		Q_{m1} &= W_q \cdot f_{m1} \\
		K_{m2} &= W_k \cdot f_{m2} \\
		V_{m2} &= W_v \cdot f_{m2}
	\end{align}
	
	\begin{figure*}[t!]
		\centering
		\includegraphics[width=0.7\textwidth]{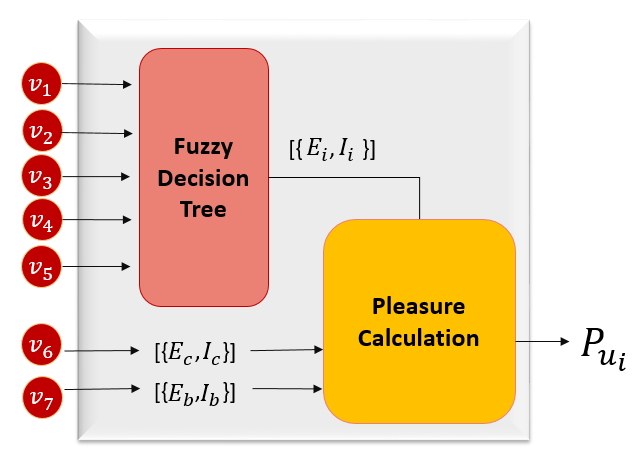}
		\caption{The Fuzzy Pleasure Inference Module (FPIM). The Fuzzy Decision Tree processes appraisal variables to generate emotion-intensity pairs $ [\{E_i, I_i\}] $. For the calm and boredom emotions, the emotion-intensity pairs are obtained via direct model prediction instead of the fuzzy decision tree.}
		\label{fig:fpim}
	\end{figure*}
	
	Here, $Q_{m1}$ denotes the query vectors derived from one modality, whereas $K_{m2}$ and $V_{m2}$ correspond to the key and value vectors extracted from the complementary modality. The input features $f_{m1}$ and $f_{m2}$ represent the raw embeddings of the first and second modalities, respectively. For instance, the visual embedding $f_v$ generates the query $q_v$, while the audio embedding $f_a$ produces the corresponding key $k_a$ and value $v_a$. The cross-attention layer serves to project and align representations across modality-specific spaces.
	
	To account for intra-modal dynamics, the fusion framework utilizes self-attention mechanisms that identify temporal correlations within each modality's post-cross-attention representation. These enriched features are then processed through a position-wise feed-forward architecture, leveraging non-linear ReLU activations to polish the shared encodings. This integrated pipeline ensures the synthesis of a comprehensive joint representation, optimized for the final emotional classification task.

	\subsubsection{Multi-Task Supervision Framework}
	
	To ensure stable and effective training, we augment each modality-specific feature extractor with dedicated fully connected heads positioned at its output stage. These auxiliary branches generate independent unimodal predictions in addition to the primary multimodal output produced by the fusion pathway. This multi-task setup allows us to supervise the model using three parallel loss terms: one for each modality-specific head and one for the fused representation.
	
	While final inference relies exclusively on the fused representation, the overall training objective is defined as the weighted summation of all three loss components. This strategy encourages the individual feature networks to develop strong, modality-aligned representations while simultaneously guiding the fusion module toward coherent cross-modal integration. By enforcing consistency across both unimodal and multimodal pathways, this auxiliary supervision acts as a powerful regularization mechanism, significantly improving convergence and generalization:
	\begin{equation} \label{eq:multi_loss}
		Loss = \sum_{m \in \{a,v,t,f\}} \alpha_m \cdot \text{loss\_fn}(y_m, target_m)
	\end{equation}
	This feature enhances the model's resilience to data imperfections and the model performance remains strong when a single modality is subject to noise or data loss, provided that redundant and sufficient information is available across the complementary modalities.

	\begin{table}[!t]
		\renewcommand{\arraystretch}{1.3}
		\caption{Experimental results expressed in degrees \cite{taverner2019towards}}
		\label{tab:emotion_mapping}
		\centering
		\begin{tabular}{|c|c|c|c|c|}
			\hline
			\textbf{Emotion} & \textbf{Pleasure} & \textbf{Arousal} & \textbf{Mean Angle} & \textbf{SD} \\
			\midrule
			Happiness & 0.90 & 0.42 & 25.09 & 19.02 \\
			Excitement & 0.76 & 0.64 & 39.97 & 10.32 \\
			Surprise & 0.31 & 0.95 & 71.63 & 26.38 \\
			Fear & -0.58 & 0.81 & 125.51 & 15.61 \\
			Anger & -0.74 & 0.66 & 138.55 & 16.90 \\
			Disgust & -0.99 & -0.04 & 182.58 & 43.65 \\
			Sadness & -0.96 & -0.27 & 196.02 & 22.48 \\
			Boredom & -0.41 & -0.91 & 245.34 & 21.41 \\
			Sleepiness & -0.11 & -0.99 & 263.59 & 15.56 \\
			Calm & 0.74 & -0.67 & 318.12 & 35.89 \\
			\hline
		\end{tabular}
	\end{table}
	
	\subsubsection{Residual-Style Modality Enrichment}
	
	Cross-attention enables rich interaction between modalities, but it risks over-alignment, where one modality’s unique characteristics are overshadowed by the dominant patterns of the other. To maintain modality identity while still benefiting from cross-modal context, we introduce a Residual-Style Modality Enrichment module that keeps the original feature stream intact and merges it with the attention-augmented version. This design ensures that discriminative, modality-specific cues are not diluted during fusion, allowing the model to reason from both raw sensory evidence and inter-modal relationships. This balanced integration leads to more robust and interpretable multimodal representations.
	
	Specifically, this model fuses the original and projected hidden states along the feature dimension and employs transformer encoders for further processing of these combined hidden states. For a given utterance $u_i$, let $H_{\mathrm{audio}_{i}} \in \mathbb{R}^{T_a \times d_a}$, $H_{\mathrm{visual}_{i}} \in \mathbb{R}^{T_v \times d_v}$, $H_{\mathrm{text}_{i}} \in \mathbb{R}^{T_t \times d_t}$ represent the original hidden states for audio, visual and text modalities, respectively, and $H'_{\mathrm{audio}_{i}}, H'_{\mathrm{visual}_{i}},  H'_{\mathrm{text}_{i}}$ denote the projected hidden states after cross-attention. The fusion process concatenates these along the feature dimension as follows:
	
	\begin{align} \label{eq:signal_restoration}
		H_{\mathrm{audio\_concat}_{i}} &= \text{Concat}\left( H_{\mathrm{audio}_{i}}, H'_{\mathrm{audio}_{i}} \right), \notag \\
		H_{\mathrm{visual\_concat}_{i}} &= \text{Concat}\left( H_{\mathrm{visual}_{i}}, H'_{\mathrm{visual}_{i}} \right), \notag \\
		H_{\mathrm{text\_concat}_{i}} &= \text{Concat}\left( H_{\mathrm{text}_{i}}, H'_{\mathrm{text}_{i}} \right)
	\end{align}
	
	where:
	$H_{\mathrm{audio}_{i}} \in \mathbb{R}^{T_a \times d_a}$ represents the original audio feature representations for utterance $u_i$, derived from the audio feature network. $H_{\mathrm{visual}_{i}} \in \mathbb{R}^{T_v \times d_v}$ represents the original visual feature representations for utterance $u_i$, derived from the visual feature network after projection and $H_{\mathrm{text}_{i}} \in \mathbb{R}^{T_t \times d_t}$ represents the original text feature representations for utterance $u_i$, derived from the textual feature network after projection. $H'_{\mathrm{audio}_{i}} \in \mathbb{R}^{T_a \times d_a}$, $H'_{\mathrm{visual}_{i}} \in \mathbb{R}^{T_v \times d_v}$ and $H'_{\mathrm{text}_{i}} \in \mathbb{R}^{T_t \times d_t}$ denote the projected hidden states after cross-modal attention in the fusion module. Concatenating the original and projected features along the feature dimension, yielding $H_{\mathrm{audio\_concat}_{i}} \in \mathbb{R}^{T_a \times 2d_a}$, $H_{\mathrm{visual\_concat}_{i}} \in \mathbb{R}^{T_v \times 2d_v}$ and $H_{\mathrm{text\_concat}_{i}} \in \mathbb{R}^{T_t \times 2d_t}$.
	
	These concatenated features are prepended with a CLS token and processed by Transformer encoders with $d_{\text{model}}$ and 12 attention heads:
	\begin{align}
		H_{\mathrm{audio\_mixed}_{i}} &= \text{TEncoder}\left( \text{CLS}\left( H_{\mathrm{audio\_concat}_{i}} \right), \text{Mask}_{\mathrm{audio}_{i}} \right) \\
		H_{\mathrm{visual\_mixed}_{i}} &= \text{TEncoder}\left( \text{CLS}\left( H_{\mathrm{visual\_concat}_{i}} \right), \text{Mask}_{\mathrm{visual}_{i}} \right) \\
		H_{\mathrm{text\_mixed}_{i}} &= \text{TEncoder}\left( \text{CLS}\left( H_{\mathrm{text\_concat}_{i}} \right), \text{Mask}_{\mathrm{text}_{i}} \right)
	\end{align}
	where TEncoder denotes the transformer encoder module. The final fused representation is:
	\begin{equation}
		H_{\mathrm{fused}_{i}} = \text{Concat}\left( H_{\mathrm{audio\_mixed}_{i}}^{[\mathrm{CLS}]}, H_{\mathrm{visual\_mixed}_{i}}^{[\mathrm{CLS}]}, H_{\mathrm{text\_mixed}_{i}}^{[\mathrm{CLS}]} \right)
	\end{equation}
	where $H^{[\mathrm{CLS}]}$ denotes the hidden state corresponding to the CLS token in the respective mixed representations, yielding $H_{\mathrm{fused}_{i}} \in \mathbb{R}^{2d_{model}}$.
	Finally, $H_{\mathrm{fused}_{i}}$ is passed through a linear layer to produce the final output.

	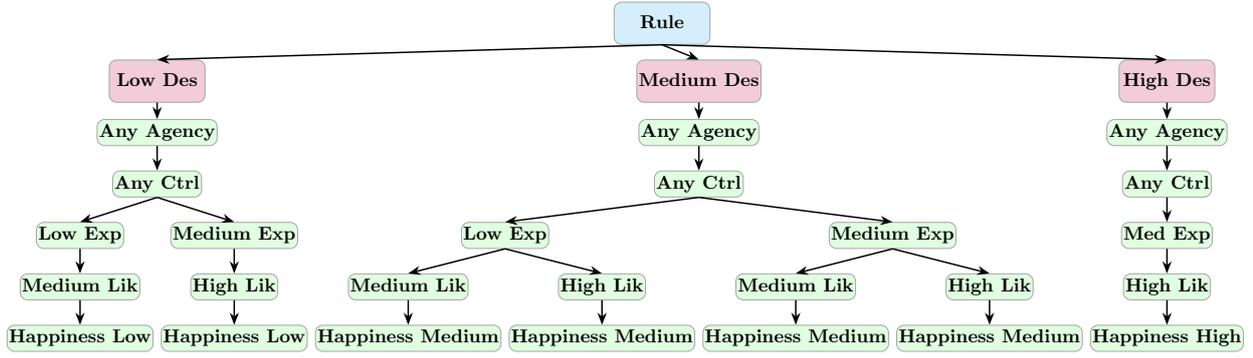
\begin{figure*}[!t]
		\centering
		\caption{Appraisal-driven Fuzzy Decision Tree for Emotion Recognition and Intensity Estimation}
		\label{fig:appraisal_tree_final}
		
		\makebox[\textwidth][c]{
			\scalebox{0.70}{
				\begin{forest}
					for tree={
						font=\sffamily\tiny, 
						edge={->, thick, >=Stealth},
						l sep=8pt,   
						s sep=4pt,   
						inner sep=1pt, 
						rounded corners,
						draw=gray!70,
						align=center,
						parent anchor=south,
						child anchor=north,
						if level=0{fill=cyan!15, font=\bfseries\normalsize, minimum width=1.8cm, minimum height=0.8cm}{},
						if level=1{fill=purple!20,font=\bfseries\normalsize, minimum width=1.8cm, minimum height=0.8cm}{},
						if level>=2{fill=green!12, font=\bfseries}{}
					}
					[Rule
					[Low Des
					[Any Agency
					[Any Ctrl
					[Low Exp
					[Medium Lik
					[Happiness Low]
					]
					]
					[Medium Exp
					[High Lik
					[Happiness Low]
					]
					]
					]
					]
					]
					[Medium Des
					[Any Agency
					[Any Ctrl
					[Low Exp
					[Medium Lik
					[Happiness Medium]
					]
					[High Lik
					[Happiness Medium]
					]
					]
					[Medium Exp
					[Medium Lik
					[Happiness Medium]
					]
					[High Lik
					[Happiness Medium]
					]
					]
					]
					]
					]
					[High Des
					[Any Agency
					[Any Ctrl
					[Med Exp
					[High Lik
					[Happiness High]
					]
					]
					]
					]
					]
					]
			\end{forest}}
		}
		\vspace{14pt}
		
		\makebox[\textwidth][c]{
			\scalebox{0.70}{
				\begin{forest}
					for tree={
						font=\sffamily\tiny, 
						edge={->, thick, >=Stealth},
						l sep=8pt,   
						s sep=4pt,   
						inner sep=1pt, 
						rounded corners,
						draw=gray!70,
						align=center,
						parent anchor=south,
						child anchor=north,
						if level=0{fill=cyan!15, font=\bfseries\normalsize, minimum width=1.8cm, minimum height=0.8cm}{},
						if level=1{fill=purple!20,font=\bfseries\normalsize, minimum width=1.8cm, minimum height=0.8cm}{},
						if level>=2{fill=green!12, font=\bfseries}{}
					}
					[Rule
					[High Undes
					[Other Agency
					[Low Ctrl
					[Low Exp
					[Low Lik
					[Surprise High]
					]
					[Medium Lik
					[Surprise/Fear Medium]
					]
					[High Lik
					[Surprise Low/Fear High]
					]
					]
					]
					[Medium/High Ctrl
					[Low Exp
					[High Lik
					[Disgust High]
					]
					]
					]
					]
					[None Agency
					[Low Ctrl
					[Any Exp
					[Low Lik
					[Sadness Low]
					]
					[Medium Lik
					[Sadness Medium]
					]
					[High Lik
					[Sadness High]
					]
					]
					]
					[Medium Ctrl
					[Medium Exp
					[High Lik
					[Anger High]
					]
					]
					]
					]
					]
					]
			\end{forest}}
		}
		\vspace{14pt}
		
		\makebox[\textwidth][c]{
			\scalebox{0.70}{
				\begin{forest}
					for tree={
						font=\sffamily\tiny,
						edge={->, thick, >=Stealth},
						l sep=8pt,
						s sep=4pt,
						inner sep=1.5pt,
						rounded corners,
						draw=gray!70,
						align=center,
						parent anchor=south,
						child anchor=north,
						if level=0{fill=cyan!15, font=\bfseries\normalsize, minimum width=1.8cm, minimum height=0.8cm}{},
						if level=1{fill=purple!20, font=\bfseries\normalsize, minimum width=1.8cm, minimum height=0.8cm}{},
						if level>=2{fill=green!12, font=\bfseries}{}
					}
					[Rule
					[Low Undes
					[Other Agency
					[Low Ctrl
					[Low Exp
					[High/Medium/Low Lik
					[Fear Low]
					]
					]
					]
					[High/Medium Ctrl
					[Low Exp
					[High Lik
					[Disgust Low]
					]
					]
					]
					]
					[None Agency
					[Low Ctrl
					[High/Medium/Low Lik
					[Sadness Low]
					]
					]
					[Medium Ctrl
					[Medium Exp
					[High/Medium/Low Lik
					[Anger Low]
					]
					]
					]
					]
					]
					[Medium Undes
					[Other Agency
					[High/Medium Ctrl
					[Low Exp
					[High Lik
					[Disgust Medium]
					]
					]
					]
					]
					[None Agency
					[Low Ctrl
					[Any Exp
					[Medium Lik
					[Sadness Medium]
					]
					[High Lik
					[High/Medium Sadness]
					]
					]
					]
					[Medium Ctrl
					[Medium Exp
					[High/Medium Lik
					[Anger Medium]
					]
					]
					]
					]
					]
					]
					]
			\end{forest}}
		}
		\vspace{10pt}
		\textbf{Legend}: Exp = Expectedness, Lik = Likelihood,\\
		Des = Desirable, Undes = Undesirable, Ctrl = Controllability
		\scriptsize
	\end{figure*}
	
	\subsection{The Fuzzy Pleasure Inference Module (FPIM)}
	The FPIM submodule, represented by a function $\mathcal{F}$, takes the predicted intermediate variables $V_i$ as input to compute the final Pleasure value $P_i$. This relationship is defined as:
	\begin{equation}
		P_i = \mathcal{F}(V_i)
	\end{equation}
	Following the prediction of appraisal variables by the CAFM, we leverage a fuzzy decision tree to quantify the resulting pleasure. This module generates both a numerical intensity value and a linguistic label to represent the induced pleasure. This system evaluates various emotional states and their intensities, with the process detailed in subsequent sections. Figure \ref{fig:fpim} illustrates the specifics of this module.
	
	This section details the proposed Fuzzy Pleasure Inference Module, designed to evaluate utterances and infer emotional states within a pleasure-arousal space. This module leverages fuzzy logic to handle the nuanced and subjective nature of human emotion.
	
	\begin{figure*}[!t] 
		\centering
		\includegraphics[width=0.7\textwidth]{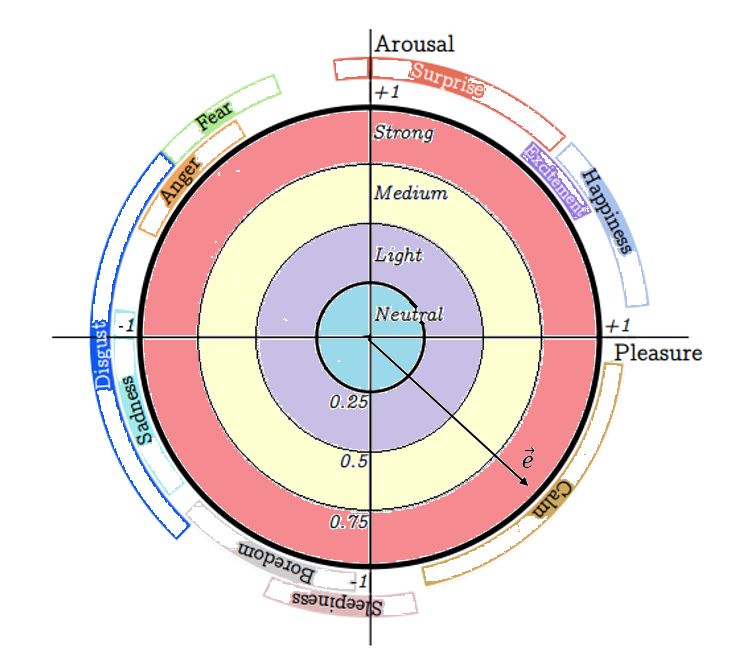}
		\caption{Example of Emotion-Pleasure Mapping}
		\label{fig:mapping}
	\end{figure*}

	\subsubsection{Fuzzy Decision Tree for Utterance Emotion Elicitation}
	
	We employ a fuzzy decision tree to map observed utterances onto discrete affective states. This structure explicitly encodes fuzzy inference rules as a hierarchical decision process, enabling interpretable and efficient emotion reasoning. Inspired by the OCC model \cite{ortony2022cognitive}, we integrated fuzzy logic into a decision tree framework to operationalize these cognitive rules into a computational format. This allows the system to derive specific emotions and their intensities by traversing the tree based on the values of the appraisal variables. The system recognizes eight primary emotion classes: Ekman’s canonical six (happiness, sadness, anger, fear, disgust, surprise) plus calm and boredom, consistent with Russell’s dimensional framework. By traversing the decision tree according to linguistic and contextual appraisal dimensions (e.g., desirability, expectedness, controllability), the model simultaneously predicts both the dominant emotion label and its graded intensity.
	
	Each utterance $u_i$ is characterized by a set of cognitive attributes:
	\begin{equation} \label{eq:utterance_attributes}
		V_{u_i} = (V_e, V_l, V_d, V_{co}, V_a)
	\end{equation}
	where $V_e$ (Expectedness), $V_l$ (Likelihood), $V_d$ (Desirability), $V_a$ (Agency), and $V_{co}$ (Controllability) are the appraisal variables. These variables take on the following values:
	\begin{itemize}
		\item  ($V_e$), ($V_l$), and ($V_{co}$): Low, Medium, High
		\item  ($V_d$): Highly Desirable, Desirable, Low Desirable, Low Undesirable, Undesirable, Highly Undesirable
		\item  ($V_a$): Other
	\end{itemize}
	
	Each emotion is defined by its type and intensity, with possible values as follows:
	\begin{itemize}
		\item Emotion Type: Happiness, Sadness, Anger, Fear, Disgust, Surprise, Calm, Boredom
		\item Intensity: High, Medium, Low
	\end{itemize}
	
	Our fuzzy decision tree is employed to compute the evoked emotions and their intensity. The system organizes its fuzzy inference logic as a hierarchical decision process: appraisal variables are evaluated sequentially along tree branches, with each path encoding a specific combination of conditions that leads to a unique emotional outcome. The systematic procedure for this hierarchical  evaluation is detailed in algorithm \ref{alg:tree}. Formally, each leaf node corresponds to a rule $r_i$ such that, if Expectedness is $x_E^i$, Likelihood is $x_L^i$, Desirability is $x_D^i$, Agency is $x_A^i$, and Controllability is $x_C^i$, then the emotion is $y_{\text{emotion}}^i$ with an intensity of $y_{\text{intensity}}^i$. For example, one path in the tree encodes: “If Desirability is Highly Desirable, Expectedness is High, and Likelihood is High, then the emotion is Happiness with High intensity.” In total, the tree implements 33 distinct appraisal-to-emotion mappings, as illustrated in Figure \ref{fig:appraisal_tree_final}, each corresponding to a unique leaf that specifies a configuration of appraisal variables and its associated emotional label and intensity.

	\begin{algorithm}[!t]
		\caption{Inference Procedure for the Appraisal-Driven Fuzzy Decision Tree}
		\label{alg:tree}
		\begin{algorithmic}[1] 
			\Require Appraisal vector $V = \{v_e, \dots, v_b\}$
			\Ensure Predicted emotion type and intensity level
			\State Start at the root node (Desirability)		
			\While{current node is not a leaf node}		
			\State Let $A_i$ be the appraisal variable of the current node
			\State Let $S_i$ be the set of allowed fuzzy linguistic values
			\State Determine the active child node:
			\State \hspace{1em} Find child whose condition best matches $v_i \in V$
			\State \hspace{1em} ($v_i$ falls into a linguistic term in $S_i$)
			\State \hspace{1em} In case of overlap (e.g., Low/Medium), follow the defined rule path
			\State Move to the selected child node	
			\EndWhile
			\State \Return Emotion type and intensity from the leaf node
		\end{algorithmic}
	\end{algorithm}
	
	\subsubsection{Mapping Emotions to Pleasure-Arousal Space}
	
	The fuzzy decision tree produces a fuzzy affective outcome, consisting of a primary emotion category and its associated intensity level. This fuzzy-labeled emotion is subsequently projected into a continuous pleasure-arousal coordinate system. The mapping is grounded in empirical data from \cite{taverner2019towards}, where human raters provided subjective pleasure and arousal scores for emotion labels under uncertainty. From these responses, the mean angles and standard deviations were derived for each emotion in the 2D affective space. These statistics are compiled in Table \ref{tab:emotion_mapping} to guide the placement of fuzzy emotion outputs.
	
	The pleasure-arousal space is also divided into four intensity degrees: Strong, Medium, Light, and Neutral as illustrated in figure \ref{fig:mapping}. The neutral fuzzy value is excluded from our model, as it refers to emotions not intense enough to be considered evoked.
	\subsubsection{Defuzzification to Pleasure Value}
	
	A defuzzification module transforms the fuzzy emotional appraisal into a crisp pleasure score by interpreting it within the 2D pleasure-arousal plane. It first assigns a directional angle to the inferred emotion vector $e$, drawn from the experimentally derived mean angles (see Table \ref{tab:emotion_mapping}). For instance, an appraised emotion of Calm is aligned at 318.12°, reflecting the average angular position reported in \cite{taverner2019towards}.
	
	The magnitude of encoding emotional intensity $e$is then computed as the predefined scalar value associated with the fuzzy intensity label (e.g., High, Medium, Low). These intensity magnitudes are set as fixed averages calibrated to reflect perceived strength across emotion categories. This vector-based representation enables precise, continuous pleasure estimation while preserving the graded nature of the fuzzy inference output.
	
	Given that the horizontal axis in the two-dimensional space represents pleasure and the vertical axis represents arousal, if $\alpha$ is the angle of vector $e$, the pleasure value of the appraised emotion can be calculated using a simple trigonometric formula:
	\begin{equation} \label{eq:pleasure_alpha}
		\text{Pleasure} = |e| \cdot \cos \alpha
	\end{equation}
	
	Since emotions like Calm and Boredom can be evoked simultaneously with other, even conflicting emotions, the final pleasure value is calculated as the average of the pleasure values of the activated emotions from the fuzzy decision tree:
	\begin{equation} \label{eq:final_pleasure}
		\text{Pleasure}_{\text{final}} = \frac{\sum_{i=1}^{n} w_i \cdot \text{Pleasure}_i}{\sum_{i=1}^{n} w_i}
	\end{equation}
	where $n$ is the number of activated fuzzy paths corresponding to different emotions (i.e., the number of distinct emotions contributing to the pleasure calculation), $w_i$ is the weight of emotion $i$ in the pleasure calculation, equivalent to the number of conditions that activate the rule corresponding to that emotion. For instance, if three conditions are required to activate the rule for Happiness and these three conditions are met, Happiness will be included in the pleasure calculation with a weight of 3.
	
	For test data, the Pleasure value is inferred by first using the trained model $\mathcal{M}$ to predict the appraisal variables and then using the fuzzy model $\mathcal{F}$ to compute the final Pleasure value.
	
	\section{Experimental Setup}\label{sec:experiments}
	
	This section provides a comprehensive evaluation of the proposed multimodal pleasure computation model. It details the experimental setup, evaluation metrics, preprocessing steps, and the quantitative results obtained from both the Cognitive Appraisal Fusion Module (CAFM) and the Fuzzy Pleasure Inference Module (FPIM). Analyses of various optimizations, including the impact of appraisal variable mapping, increased image model parameters, and extended audio sampling rates, are presented.
	
	\subsection{Feature Extraction}
	
	This section details the specific models and methodologies used for extracting features from the audio and visual modalities. The preprocessing steps for each modality are described in the Evaluation Setup subsection.
	\subsubsection{Audio Extraction}
	
	Audio data was preprocessed to ensure a consistent format. Subsequently, \texttt{wav2vec} \cite{baevski2020wav2vec}, a model pre-trained on large-scale audio datasets, is utilized to extract foundational audio patterns, processing raw waveforms into feature vectors that capture temporal and spectral characteristics. These initial features are then fed into the \texttt{Data2Vec} \cite{baevski2022data2vec} model, a self-supervised architecture, which further enriches them to represent emotional aspects such as tonality and rhythm, thus acquiring dynamic audio information crucial for pleasure analysis.
	
	\subsubsection{Video Extraction}
	
	For visual data, video frames are initially extracted from raw data and subjected to necessary preprocessing. Following this, the \texttt{TimeSformer} \cite{bertasius2021space} model extracts meaningful features from these frames. \texttt{TimeSformer} extends the Vision Transformer (ViT) \cite{dosovitskiy2020image} architecture to effectively handle spatio-temporal data. This Transformer processes video frames by partitioning them into patches and applying self-attention mechanisms across both spatial and temporal dimensions. Collectively, this advanced feature extraction methodology establishes a robust foundation for the subsequent stages of multimodal fusion and appraisal prediction.

	\subsection{Evaluation Setup}
	
	This subsection outlines the detailed configurations and structures used for evaluating the proposed method. These settings encompass the dataset utilized, evaluation metrics, and specific configurations for model training. The objective is to precisely assess the model's ability to compute video-induced pleasure.
	
	\subsubsection{Dataset}
	In this study, the EmoStim dataset \cite{somarathna2023emostim} was used, comprising 99 emotionally annotated video clips. Each clip features multi-participant ratings across discrete emotion categories and 39 appraisal-related dimensions grounded in Scherer’s Component Process Model (CPM). These dimensions encompass appraisal, motivation, expression, physiology, and subjective feeling. In our work, seven of these features were selected to drive the Fuzzy Pleasure Inference Module (FPIM). To enhance computational efficiency and performance, video clips were segmented into shorter intervals with a maximum duration of 10 seconds. Since the EmoStim dataset lacked explicit labels for pleasure (Pleasant, Unpleasant, and Neutral), a supplementary annotation phase was conducted. Thirty participants (13 females, 17 males, $M_{age} = 31$) were recruited, with each participant randomly assigned 10 videos. This ensured that each clip received three independent ratings, with final labels determined via majority vote. Participants were instructed to report their genuine emotional experiences rather than perceived social norms. Each video was labeled using both two-class and three-class classification schemes to facilitate a comprehensive evaluation of the model. While the proposed framework is inherently multimodal, the textual modality was excluded from the current implementation due to its unavailability in the utilized dataset. However, the architecture is designed to be extensible, allowing for the seamless integration of textual features in future studies.
	
	\subsubsection{Data Split}
	The EmoStim dataset was divided into three distinct sets for model training, validation, and evaluation. The training set comprised 938 samples, used for model training. The validation set, with 264 samples, was designed to tune hyperparameters and prevent overfitting. Finally, the test set, consisting of 157 samples, was used for the final performance evaluation of the model. This division followed an approximate ratio of 70\% (training), 20\% (validation), and 10\% (testing), ensuring a balanced approach to model generalization.

	\subsubsection{Evaluation Metrics}
	The proposed model's performance is assessed across two main stages: training and inference stage. During the training phase, the model is developed as a regression problem, predicting continuous values for emotional appraisal variables such as desirability, expectedness, likelihood, controllability, calm, and boredom. This regression approach enables the model to precisely compute complex relationships between multimodal audio and visual data, providing numerical predictions ranging from 0 to 5.
	
	For optimization and performance evaluation at this stage, the ACC3 metric is employed to assess prediction accuracy across discretized levels. This metric is defined as:
	\begin{equation}ACC3 = \frac{1}{N} \sum_{i=1}^{N} I (\mathbb{F}(\hat{y}_i) = \mathbb{F}(y_i))\end{equation}

	where $\mathbb{F}(\cdot)$ is an adaptive binning function that maps both the predicted values ($\hat{y}_i$) and the ground truth labels ($y_i$) into three categories: High, Medium, and Low. This function is adaptive, utilizing either fixed-interval thresholds or clustering-based boundaries as detailed in the following section. $I(\cdot)$ represents the indicator function, which returns $1$ if the predicted and actual categories match, and $0$ otherwise. Finally, $N$ denotes the total number of samples, with the resulting ACC3 value representing the overall percentage of correctly categorized instances.
	
	Conversely, the final evaluation stage focuses on a classification problem specifically designed for pleasure computation and analysis. After the model predicts appraisal variables, pleasure values are numerically calculated and then mapped to linguistic labels at two levels (Pleasant and Unpleasant) and three levels (Pleasant, Unpleasant, and Neutral). To assess the accuracy of these classifications, a comprehensive set of metrics such as Accuracy, Precision, Recall, F1-Score is utilized. The model's performance on imbalanced datasets was comprehensively evaluated using Macro-Precision, Macro-Recall, Weighted-Precision, and Weighted-Recall.
	
	\subsubsection{Preprocessing}
	To ensure the quality of input data and the effectiveness of feature extraction, preprocessing steps were applied to both visual and audio data.
	
	\textbf{Visual Preprocessing:} Video frames were converted from BGR to RGB format for model compatibility. Frames were resized to 224x224 pixels (in initial model) and 448x448 pixels (in the second model), and normalized using a mean of [0.485, 0.456, 0.406] and a standard deviation of [0.229, 0.224, 0.225] for each color channel to ensure uniform feature distribution. Eight frames (in initial model) and 16 frames (in the second model) were selected using linear sampling to represent the entire video. In cases of insufficient frames, zero-padding was applied to maintain a consistent number.
	
	\textbf{Audio Preprocessing:} Audio features were extracted using the Wav2Vec2 \cite{baevski2020wav2vec} extractor, with the following steps: Stereo audio signals were converted to a mono channel by averaging the channels to simplify processing. Audio data was then normalized using the extractor's settings to standardize its amplitude. Signals were either truncated to a maximum length of 96000 samples (6 seconds) in initial model and 160000 samples (10 seconds) in the second model, or padded to a fixed length to meet model requirements. Features were extracted at a sampling rate of 16000 Hz, employing an attention mask to identify invalid or padded sections of the signal.
	
	\subsubsection{Model Training and Parameters}
	The model was trained using the AdamW optimizer \cite{loshchilov2017decoupled}. The L1Loss function was employed to compute the absolute difference between predictions and actual values. The model was trained for 20 epochs with a learning rate of 1e-5 and a batch size of 2. An early stopping mechanism with a patience of 8 epochs was utilized to prevent overfitting, and a dropout rate of 0.3 was applied to enhance model generalization. Also a random seed of zero was set for reproducibility. For the audio pre-trained model, the Convolutional Neural Network (CNN) portion used for feature extraction was frozen.

In the following section, we present the quantitative results obtained from the proposed framework.
	\section{Results}
	
	This section reports the experimental results, including data-driven appraisal prediction accuracy and theory-driven pleasure inference performance. Since this study represents a pioneering effort in integrating appraisal-driven fuzzy decision trees for pleasure estimation, direct comparative benchmarks with existing literature are currently unavailable. Consequently, the performance is evaluated against the established ground truth of the annotated dataset to provide a baseline for this novel approach. First, we examine the outcomes from the Cognitive Appraisal Fusion Module (CAFM), followed by an interpretation of the results obtained from the Fuzzy Pleasure Inference Module (FPIM).
	
	\subsection{Data-Driven Evaluation}
	
	Initially, the model's accuracy was calculated using fixed ranges for prediction classification. Subsequently, a clustering-based approach was implemented to determine classification ranges, leading to optimized model performance. Further enhancements were achieved by applying modifications to audio and visual feature extraction, as detailed below.
	
	\textbf{Initial Accuracy with Fixed Ranges:} In the first stage, the model's accuracy was computed using fixed ranges for prediction classification. For binary classification, predicted and true values were divided based on a threshold of 3, meaning values less than 3 were assigned to one class, and values equal to or greater than 3 to another. This method calculated binary accuracy, reflecting the model's ability to distinguish between low and high values. For three-class classification, two distinct approaches were applied: a soft version, where values were divided into three groups (less than 2.5, between 2.5 and 3.5, and greater than 3.5) offering greater classification flexibility; and a strict version, with divisions based on values up to 2, equal to 3, and greater than 3, imposing more stringent constraints. These fixed ranges served as a starting point for the model's initial evaluation, providing a baseline for subsequent comparisons.
	
	\textbf{Optimization of Ranges with Clustering:} In the next stage, a clustering-based approach was utilized to optimize model performance by determining dynamic classification ranges. For each appraisal variable, the k-means clustering algorithm was applied to the data to identify three natural clusters. Based on the boundary points derived from these clusters, new ranges for three-class classification were defined. For example, for variable Desirability, the ranges were set based on values of 1.72 and 3.44, such that values up to 1.72 fell into the first cluster, between 1.72 and 3.44 into the second, and above 3.44 into the third. This process was repeated for 7 key variables, with each variable's specific ranges adjusted according to its data distribution. The results demonstrated that this method significantly improved accuracy; for instance, the three-class accuracy for some variables, such as Desirablity and Likelihood, notably improved, indicating better alignment of the ranges with the data's intrinsic structure. Table \ref{tab:kmeans_boundaries} reports the boundaries obtained from clustering for each variable.
	\begin{center}
		\captionof{table}{Boundaries obtained from K-means Clustering for Appraisal Variables}
		\label{tab:kmeans_boundaries}
		\begin{minipage}{0.8\textwidth}
			\centering
			\renewcommand{\arraystretch}{1.3}
			\begin{tabular}{|l|c|}   
				\hline
				\textbf{Appraisal Variable} & \textbf{K-means Boundaries} \\
				\hline
				Desirability    & [1.72, 3.44] \\
				Calm            & [1.72, 3.47] \\
				Boredom         & [1.69, 3.50] \\
				Controllability & [1.71, 3.34] \\
				Likelihood      & [1.75, 3.43] \\
				Expectedness    & [1.75, 3.37] \\
				\hline
			\end{tabular}
		\end{minipage}
	\end{center}
	The results from these stages are summarized in Table \ref{tab:initial_model_accuracy}. For simplification, the highest three-class accuracy obtained from k-means clustering is reported alongside the binary and three-class soft and strict accuracies for each variable. These results indicate significant accuracy improvement with clustering, providing a basis for further analysis. Binary accuracy is also reported in Table \ref{tab:initial_model_accuracy}, but given that predicted values range from 1 to 5 and are intended to be categorized into three bins (High, Medium, Low) for final mapping, three-class accuracy holds greater importance in this study. Therefore, model evaluation primarily focuses on three-class accuracy.
	
	\begin{center}
		\captionof{table}{Initial Model Accuracy Results for Appraisal Variables}
		\label{tab:initial_model_accuracy}
		\begin{minipage}{0.95\textwidth}
			\centering
			\renewcommand{\arraystretch}{1.3}
			\begin{tabular}{|l|c|c|c|c|}
				\hline
				\textbf{Appraisal Variable} & \textbf{ACC2} & \textbf{ACC3 (soft)} & \textbf{ACC3 (strict)} & \textbf{ACC3 (kmeans)} \\
				\hline
				Desirability & 0.6306 & 0.5414 & 0.5159 & 0.5796 \\
				Calm & 0.3121 & 0.2293 & 0.2484 & 0.2739 \\
				Boredom & 0.9921 & 0.9874 & 0.9846 & 0.9893 \\
				Controllability & 0.5924 & 0.3248 & 0.1656 & 0.5223 \\
				Likelihood & 0.4013 & 0.5159 & 0.4904 & 0.6752 \\
				Expectedness & 0.4331 & 0.3885 & 0.3949 & 0.5669 \\
				\hline
			\end{tabular}
		\end{minipage}
	\end{center}
	\textbf{Impact of Increasing Visual Model Parameters and Audio Sampling Rate:}
	A key optimization step involved modifications to audio and visual feature extraction, leading to improved model performance. These changes included using a TimeSformer visual model with more parameters and an increased audio sampling rate, implemented in the second model iteration.
	
	\begin{itemize}
		\item \textbf{Increased Viusal Model Parameters:} In initial models, video frames were processed at 224x224 pixels, and 8 frames were sampled per video. In the second model, this approach was enhanced by increasing input dimensions to 448x448 pixels and the number of sampled frames to 16. These changes enabled the model to capture more visual detail and incorporate more temporal information from videos. The increase in accuracy for variables like Desirability (from 0.5796 to 0.7006) and Controllability (from 0.5223 to 0.8089) demonstrates the positive impact of increased visual feature extraction model parameters. Results indicate that this change significantly improved three-class accuracy in some variables due to the model's ability to capture more detail with additional frames and higher dimensions. However, a decrease in accuracy was observed for variables such as Calm and Expectedness, which may be attributed to data heterogeneity or the need for further model tuning.
		
		\item \textbf{Increased Audio Sampling Length:} In initial model, audio features were processed using the Wav2Vec2 extractor with a maximum length of 96000 samples (equivalent to approximately 6 seconds at a 16000 Hz sampling rate). In the second model, this maximum length was increased to 160000 samples (approximately 10 seconds). This change allowed the model to incorporate more temporal information from the audio signal, which could aid in detecting more complex emotional patterns. Table \ref{tab:second_model_accuracy} reports the results of this change for the multimodal setup. For instance, the increase in accuracy for Desirability and Controllability can be attributed to the fusion of longer audio information with visual features.
	\end{itemize}
	
	\begin{center}
		\captionof{table}{Model Accuracy Results in the Second Iteration (Impact of Increased Parameters)}
		\label{tab:second_model_accuracy}
		\begin{minipage}{0.9\textwidth}
			\centering
			\renewcommand{\arraystretch}{1.3}
			\begin{tabular}{|l|c|c|}
				\hline
				\textbf{Appraisal Variable} & \textbf{ACC3 (old)} & \textbf{ACC3 (new)} \\
				\hline
				Desirability & 0.5796 & 0.7006 \\
				Calm & 0.2739 & 0.1847 \\
				Boredom & 0.9893 & 0.9893 \\
				Controllability & 0.5223 & 0.8089 \\
				Likelihood & 0.6752 & 0.6688 \\
				Expectedness & 0.5669 & 0.3694 \\
				\hline
			\end{tabular}
		\end{minipage}
	\end{center}
	
	\subsection{Theory-Driven Evaluation (Fuzzy Pleasure Inference)}
	
	After model training, its performance in predicting pleasure labels was evaluated. The model, by fusing audio and visual features, generated predictions for emotional appraisal variables that are incorporated into the pleasure computation module. In the prediction phase, 157 samples from the test set were processed, including 157 2-class labels (Pleasant and Unpleasant) and 157 3-class labels (Pleasant, Unpleasant, and Neutral). For pleasure calculation, predicted values were mapped to fuzzy model variables, which, based on fuzzy decision tree and emotional angles (e.g., Excitement with 39.97 degrees and Disgust with 182.58 degrees) and various intensities (High, Medium, Low), converted pleasure values into linguistic labels. This process involved mapping numerical values to linguistic categories (e.g., desirability from highly undesirable to highly desirable) and calculating a weighted average, considering additional variables like calm and boredom.
	
	\textbf{2-Class Evaluation:} Results for the binary (2-class) pleasure categorization showed that the proposed model attained an overall accuracy of 0.6624, reflecting reasonable discriminative ability between pleasant and unpleasant categories. Precision was 0.6494 for the pleasant class and 0.6750 for the unpleasant class, with corresponding recall values of 0.6579 and 0.6667, respectively. The F1-scores indicate an acceptable trade-off between precision and recall for both classes. Macro- and weighted-averaged metrics were both approximately 0.66, demonstrating balanced performance across categories. Nevertheless, the achieved accuracy level may remain insufficient for applications demanding high robustness and reliability. Table \ref{tab:2class_pleasure_accuracy} reports the results of various metrics for both classes and overall accuracy.
	
	\begin{center}
		\captionof{table}{2-Class Pleasure Computation Model Accuracy Results}
		\label{tab:2class_pleasure_accuracy}
		\renewcommand{\arraystretch}{1.3}
		\begin{tabular}{|l|c|c|c|c|}
			\hline
			\textbf{Eval Metric} & \textbf{Pleasantness} & \textbf{Unpleasantness} & \textbf{Macro} & \textbf{Weighted} \\
			\hline
			Precision & 0.6494 & 0.6750 & 0.6622 & 0.6626 \\
			Recall & 0.6579 & 0.6667 & 0.6623 & 0.6624 \\
			F1 Score & 0.6536 & 0.6708 & 0.6622 & 0.6625 \\
			Support & 76 & 81 & -- & -- \\
			\hline
			\multicolumn{4}{|l|}{\textbf{Accuracy}} & 0.6624 \\
			\hline
		\end{tabular}
	\end{center}
	\textbf{3-Class Evaluation:} In the 3-class setting, the model's overall accuracy decreased to 0.5732, underscoring the greater difficulty in separating the three affective categories compared to the binary case. Precision was 0.5676 for pleasant, 0.5750 for unpleasant, and 0.6667 for neutral, while recall reached 0.5526, 0.5974, and 0.5000 for pleasant, unpleasant, and neutral classes, respectively. The F1-score for the neutral class (0.5714) indicates a reasonable trade-off between precision and recall, though performance remains constrained by class imbalance and intra-class variability in the dataset. Table \ref{tab:3class_pleasure_accuracy} reports the results of 3-class accuracy.
		
	\begin{center}
		\captionof{table}{3-Class Pleasure Computation Model Accuracy Results}
		\label{tab:3class_pleasure_accuracy}
		\renewcommand{\arraystretch}{1.3}
		\begin{tabular}{|l|c|c|c|c|c|}
			\hline
			\textbf{Eval Metric} & \textbf{Pleasantness} & \textbf{Unpleasantness} & \textbf{Neutral} & \textbf{Macro} & \textbf{Weighted} \\
			\hline
			Precision & 0.5676 & 0.5750 & 0.6667 & 0.6031 & 0.5737 \\
			Recall & 0.5526 & 0.5974 & 0.5000 & 0.5500 & 0.5732 \\
			F1 Score & 0.5600 & 0.5860 & 0.5714 & 0.5725 & 0.5730 \\
			Support & 76 & 77 & 4 & -- & -- \\
			\hline
			\multicolumn{5}{|l|}{\textbf{Accuracy}} & 0.5732 \\
			\hline
		\end{tabular}
	\end{center}

\begin{figure*}[!t]
    \centering
    \subfloat[Confusion Matrix for 2-Class Evaluation.\label{fig:confusion_matrix_2class}]{
      \includegraphics[width=0.48\textwidth]{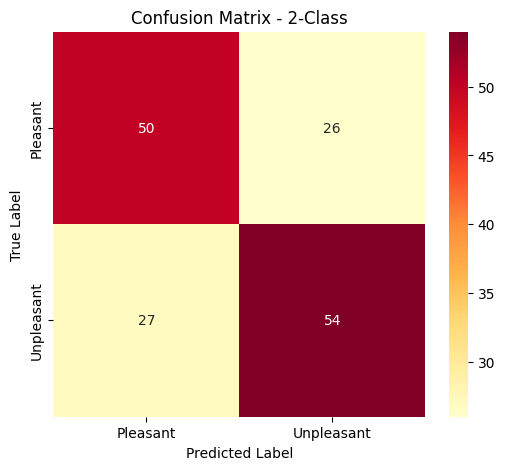}
    }
    \hfill
    \subfloat[Confusion Matrix for 3-Class Evaluation.\label{fig:confusion_matrix_3class}]{
      \includegraphics[width=0.48\textwidth]{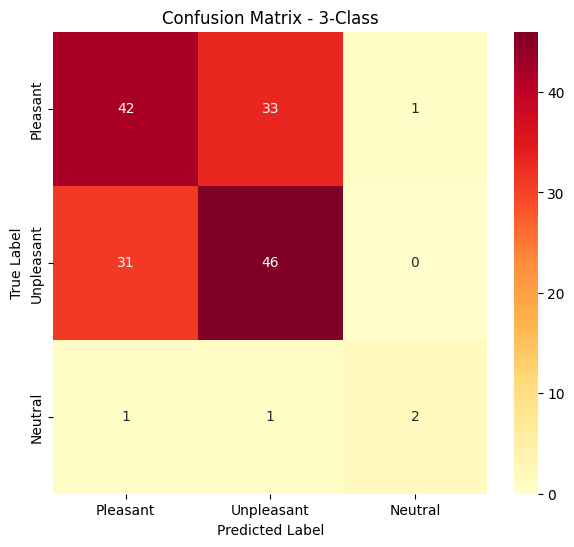}
    }
    \caption{Comparison of Confusion Matrices for 2-Class and 3-Class Evaluations.}
    \label{fig:confusion_matrices_combined}
    \end{figure*}

	\textbf{Confusion Matrices:} The confusion matrix analysis for the 3-class setting (Figure \ref{fig:confusion_matrix_3class}) reveals that 42 pleasant, 46 unpleasant, and 2 neutral instances were correctly identified. Misclassifications included 33 pleasant samples and 31 unpleasant samples assigned to incorrect categories, along with both neutral samples being erroneously predicted (one as pleasant and one as unpleasant). For the 2-class evaluation (Figure \ref{fig:confusion_matrix_2class}), the confusion matrix indicates successful classification of 50 pleasant and 54 unpleasant samples, with only 26 pleasant and 27 unpleasant samples incorrectly categorized, reflecting reasonable overall performance in binary discrimination.
	
\section{Discussion}\label{sec12}

The results of our experiments demonstrate that the proposed CogniPleasure framework, which integrates cognitive appraisal theory with multimodal feature fusion and fuzzy logic-based inference, achieves reasonable performance in predicting video-induced pleasure. In the 2-class evaluation (Pleasant vs. Unpleasant), the model attained an overall accuracy of 0.6624, with balanced precision and recall values around 0.65–0.67 across both classes. In the more challenging 3-class setting (Pleasant, Unpleasant, Neutral), the accuracy decreased to 0.5732, reflecting the greater difficulty in distinguishing subtle affective boundaries, particularly for the underrepresented neutral category.
The primary advantage of the proposed approach lies in its theory-grounded and interpretable nature. By predicting specific appraisal variables (such as desirability, controllability, likelihood, expectedness, agency) and then mapping them through a fuzzy decision tree, the model offers a causal explanation of how pleasure emerges from cognitive evaluations of video content rather than direct stimulus-response mapping. This enhances explainability in emotion detection from videos, addressing one of the key limitations of black-box deep learning models commonly used in affective computing.
The use of fuzzy logic further contributes to handling the inherent ambiguity, subjectivity, and continuous nature of human emotions. It enables a more nuanced representation of pleasure's intensity and quality (e.g., "slightly pleasurable" to "very pleasurable") and mitigates the impact of data scarcity and label noise by allowing inference even in the absence of exact rule matches. These properties make the framework particularly suitable for ecologically valid and interpretable pleasure computation.
However, several limitations remain. For emotions such as calm and boredom, model predictions were used directly because the relevant cognitive appraisal variables for these states were not explicitly defined in the existing datasets. Similarly, the neutral category was disregarded in the decision tree due to its absence in the reported data. The final pleasure score relies on a simple weighted average, which, while straightforward, may not fully capture complex interactions among appraisal variables. In addition, the current implementation excludes the textual modality due to dataset constraints, although the architecture is designed to be extensible for incorporating additional modalities such as text or physiological signals.
These limitations highlight opportunities for future improvements. The fuzzy decision tree is flexible and scalable, and can be enhanced by incorporating more comprehensive appraisal variables or by integrating emerging cognitive theories to better capture a wider spectrum of affective states, including neutral and low-activation emotions. Replacing the simple weighted average with more sophisticated mechanisms, such as using fuzzy logic itself to assign theoretically grounded weights to appraisal variables, could improve accuracy and granularity. Furthermore, advanced feature extraction and fusion methods, including large language models for textual or multimodal data, can be explored to strengthen performance while preserving interpretability. 

\section{Conclusion}\label{sec:conclusion}

In this research, we presented CogniPleasure, an innovative and interpretable framework for computing video-induced pleasure. Built upon cognitive appraisal theory, the model predicts appraisal variables directly linked to emotional experience and employs fuzzy logic with tree-structured reasoning to derive a quantifiable pleasure measure. This approach addresses key challenges in affective computing by providing a causal, theory-grounded explanation of pleasure elicitation and by effectively handling the ambiguity and subjectivity of human emotions through fuzzy inference.
Despite existing limitations, such as the direct use of predictions for calm and boredom, exclusion of neutral states, and reliance on a simple weighted average, the proposed model achieves promising results and serves as a valuable foundation for future work. By combining cognitive appraisal mechanisms with fuzzy logic, this framework paves the way for the development of more complex, explainable, and human-centric emotion detection systems from video content.	
\section*{Declerations}
\begin{itemize}
\item Funding :
The authors did not receive support from any organization for the submitted work. No funding was received to assist with the preparation of this manuscript. No funding was received for conducting this study. No funds, grants, or other support were received.
\item Conflict of interest/Competing interests :
The authors have no relevant financial or non-financial interests to disclose. The authors have no conflicts of interest to declare that are relevant to the content of this article. All authors certify that they have no affiliations with or involvement in any organization or entity with any financial interest or non-financial interest in the subject or materials discussed in this manuscript. The authors have no financial or proprietary interests in any material discussed in this article.
\item Ethics approval and consent to participate :
Not applicable
\item Consent for publication :
Not applicable
\item Data availability :

The dataset analyzed and extended during the current study is proprietary and confidential. Due to intellectual property and privacy considerations, the data are not publicly available and cannot be shared.
\item Materials availability :
Not applicable
\item Code availability :
Not applicable
\item Author contribution :

Nastaran Dab: Conceptualization, Methodology, Software, Formal analysis, Investigation, Data curation, Writing, original draft. 
Raziyeh Zall: Supervision, Validation, review \& editing. 
Mohammadreza Kangavari: Supervision, Project administration.

\end{itemize}
\bmhead{Acknowledgements}
Thanks to our research team for their continued support and helpful input.


\bibliography{sn-bibliography}


\begin{thebibliography}{88}
\ifx \bisbn   \undefined \def \bisbn  #1{ISBN #1}\fi
\ifx \binits  \undefined \def \binits#1{#1}\fi
\ifx \bauthor  \undefined \def \bauthor#1{#1}\fi
\ifx \batitle  \undefined \def \batitle#1{#1}\fi
\ifx \bjtitle  \undefined \def \bjtitle#1{#1}\fi
\ifx \bvolume  \undefined \def \bvolume#1{\textbf{#1}}\fi
\ifx \byear  \undefined \def \byear#1{#1}\fi
\ifx \bissue  \undefined \def \bissue#1{#1}\fi
\ifx \bfpage  \undefined \def \bfpage#1{#1}\fi
\ifx \blpage  \undefined \def \blpage #1{#1}\fi
\ifx \burl  \undefined \def \burl#1{\textsf{#1}}\fi
\ifx \doiurl  \undefined \def \doiurl#1{\url{https://doi.org/#1}}\fi
\ifx \betal  \undefined \def \betal{\textit{et al.}}\fi
\ifx \binstitute  \undefined \def \binstitute#1{#1}\fi
\ifx \binstitutionaled  \undefined \def \binstitutionaled#1{#1}\fi
\ifx \bctitle  \undefined \def \bctitle#1{#1}\fi
\ifx \beditor  \undefined \def \beditor#1{#1}\fi
\ifx \bpublisher  \undefined \def \bpublisher#1{#1}\fi
\ifx \bbtitle  \undefined \def \bbtitle#1{#1}\fi
\ifx \bedition  \undefined \def \bedition#1{#1}\fi
\ifx \bseriesno  \undefined \def \bseriesno#1{#1}\fi
\ifx \blocation  \undefined \def \blocation#1{#1}\fi
\ifx \bsertitle  \undefined \def \bsertitle#1{#1}\fi
\ifx \bsnm \undefined \def \bsnm#1{#1}\fi
\ifx \bsuffix \undefined \def \bsuffix#1{#1}\fi
\ifx \bparticle \undefined \def \bparticle#1{#1}\fi
\ifx \barticle \undefined \def \barticle#1{#1}\fi
\bibcommenthead
\ifx \bconfdate \undefined \def \bconfdate #1{#1}\fi
\ifx \botherref \undefined \def \botherref #1{#1}\fi
\ifx \url \undefined \def \url#1{\textsf{#1}}\fi
\ifx \bchapter \undefined \def \bchapter#1{#1}\fi
\ifx \bbook \undefined \def \bbook#1{#1}\fi
\ifx \bcomment \undefined \def \bcomment#1{#1}\fi
\ifx \oauthor \undefined \def \oauthor#1{#1}\fi
\ifx \citeauthoryear \undefined \def \citeauthoryear#1{#1}\fi
\ifx \endbibitem  \undefined \def \endbibitem {}\fi
\ifx \bconflocation  \undefined \def \bconflocation#1{#1}\fi
\ifx \arxivurl  \undefined \def \arxivurl#1{\textsf{#1}}\fi
\csname PreBibitemsHook\endcsname

\bibitem[\protect\citeauthoryear{Hanjalic and Xu}{2005}]{hanjalic2005affective}
\begin{barticle}
\bauthor{\bsnm{Hanjalic}, \binits{A.}},
\bauthor{\bsnm{Xu}, \binits{L.-Q.}}:
\batitle{Affective video content representation and modeling}.
\bjtitle{IEEE transactions on multimedia}
\bvolume{7}(\bissue{1}),
\bfpage{143}--\blpage{154}
(\byear{2005})
\end{barticle}
\endbibitem

\bibitem[\protect\citeauthoryear{Xu et~al.}{2014}]{xu2014three}
\begin{barticle}
\bauthor{\bsnm{Xu}, \binits{M.}},
\bauthor{\bsnm{Wang}, \binits{J.}},
\bauthor{\bsnm{He}, \binits{X.}},
\bauthor{\bsnm{Jin}, \binits{J.S.}},
\bauthor{\bsnm{Luo}, \binits{S.}},
\bauthor{\bsnm{Lu}, \binits{H.}}:
\batitle{A three-level framework for affective content analysis and its case
  studies}.
\bjtitle{Multimedia tools and applications}
\bvolume{70}(\bissue{2}),
\bfpage{757}--\blpage{779}
(\byear{2014})
\end{barticle}
\endbibitem

\bibitem[\protect\citeauthoryear{Canini et~al.}{2012}]{canini2012affective}
\begin{barticle}
\bauthor{\bsnm{Canini}, \binits{L.}},
\bauthor{\bsnm{Benini}, \binits{S.}},
\bauthor{\bsnm{Leonardi}, \binits{R.}}:
\batitle{Affective recommendation of movies based on selected connotative
  features}.
\bjtitle{IEEE Transactions on Circuits and Systems for Video Technology}
\bvolume{23}(\bissue{4}),
\bfpage{636}--\blpage{647}
(\byear{2012})
\end{barticle}
\endbibitem

\bibitem[\protect\citeauthoryear{Nath et~al.}{2020}]{nath2020comparative}
\begin{bchapter}
\bauthor{\bsnm{Nath}, \binits{D.}},
\bauthor{\bsnm{Anubhav}},
\bauthor{\bsnm{Singh}, \binits{M.}},
\bauthor{\bsnm{Sethia}, \binits{D.}},
\bauthor{\bsnm{Kalra}, \binits{D.}},
\bauthor{\bsnm{Indu}, \binits{S.}}:
\bctitle{A comparative study of subject-dependent and subject-independent
  strategies for eeg-based emotion recognition using lstm network}.
In: \bbtitle{Proceedings of the 2020 4th International Conference on Compute
  and Data Analysis},
pp. \bfpage{142}--\blpage{147}
(\byear{2020})
\end{bchapter}
\endbibitem

\bibitem[\protect\citeauthoryear{Zhu et~al.}{2020}]{zhu2020affective}
\begin{barticle}
\bauthor{\bsnm{Zhu}, \binits{Y.}},
\bauthor{\bsnm{Chen}, \binits{Z.}},
\bauthor{\bsnm{Wu}, \binits{F.}}:
\batitle{Affective video content analysis via multimodal deep quality embedding
  network}.
\bjtitle{IEEE Transactions on Affective Computing}
\bvolume{13}(\bissue{3}),
\bfpage{1401}--\blpage{1415}
(\byear{2020})
\end{barticle}
\endbibitem

\bibitem[\protect\citeauthoryear{Yi et~al.}{2024}]{yi2024emotion}
\begin{barticle}
\bauthor{\bsnm{Yi}, \binits{Y.}},
\bauthor{\bsnm{Zhou}, \binits{J.}},
\bauthor{\bsnm{Wang}, \binits{H.}},
\bauthor{\bsnm{Tang}, \binits{P.}},
\bauthor{\bsnm{Wang}, \binits{M.}}:
\batitle{Emotion recognition in user-generated videos with long-range
  correlation-aware network}.
\bjtitle{IET Image Processing}
\bvolume{18}(\bissue{12}),
\bfpage{3288}--\blpage{3301}
(\byear{2024})
\end{barticle}
\endbibitem

\bibitem[\protect\citeauthoryear{Berridge and
  Kringelbach}{2015}]{berridge2015pleasure}
\begin{barticle}
\bauthor{\bsnm{Berridge}, \binits{K.C.}},
\bauthor{\bsnm{Kringelbach}, \binits{M.L.}}:
\batitle{Pleasure systems in the brain}.
\bjtitle{Neuron}
\bvolume{86}(\bissue{3}),
\bfpage{646}--\blpage{664}
(\byear{2015})
\end{barticle}
\endbibitem

\bibitem[\protect\citeauthoryear{Berridge and
  Kringelbach}{2011}]{berridge2011building}
\begin{barticle}
\bauthor{\bsnm{Berridge}, \binits{K.C.}},
\bauthor{\bsnm{Kringelbach}, \binits{M.L.}}:
\batitle{Building a neuroscience of pleasure and well-being}.
\bjtitle{Psychology of Well-Being: Theory, Research and Practice}
\bvolume{1}(\bissue{1}),
\bfpage{3}
(\byear{2011})
\end{barticle}
\endbibitem

\bibitem[\protect\citeauthoryear{Moccia et~al.}{2018}]{moccia2018experience}
\begin{barticle}
\bauthor{\bsnm{Moccia}, \binits{L.}},
\bauthor{\bsnm{Mazza}, \binits{M.}},
\bauthor{\bsnm{Nicola}, \binits{M.D.}},
\bauthor{\bsnm{Janiri}, \binits{L.}}:
\batitle{The experience of pleasure: a perspective between neuroscience and
  psychoanalysis}.
\bjtitle{Frontiers in human neuroscience}
\bvolume{12},
\bfpage{359}
(\byear{2018})
\end{barticle}
\endbibitem

\bibitem[\protect\citeauthoryear{Kringelbach and
  Berridge}{2009}]{kringelbach2009towards}
\begin{barticle}
\bauthor{\bsnm{Kringelbach}, \binits{M.L.}},
\bauthor{\bsnm{Berridge}, \binits{K.C.}}:
\batitle{Towards a functional neuroanatomy of pleasure and happiness}.
\bjtitle{Trends in cognitive sciences}
\bvolume{13}(\bissue{11}),
\bfpage{479}--\blpage{487}
(\byear{2009})
\end{barticle}
\endbibitem

\bibitem[\protect\citeauthoryear{Hassouneh
  et~al.}{2020}]{hassouneh2020development}
\begin{barticle}
\bauthor{\bsnm{Hassouneh}, \binits{A.}},
\bauthor{\bsnm{Mutawa}, \binits{A.}},
\bauthor{\bsnm{Murugappan}, \binits{M.}}:
\batitle{Development of a real-time emotion recognition system using facial
  expressions and eeg based on machine learning and deep neural network
  methods}.
\bjtitle{Informatics in Medicine Unlocked}
\bvolume{20},
\bfpage{100372}
(\byear{2020})
\end{barticle}
\endbibitem

\bibitem[\protect\citeauthoryear{Li et~al.}{2022}]{li2022novel}
\begin{barticle}
\bauthor{\bsnm{Li}, \binits{R.}},
\bauthor{\bsnm{Ren}, \binits{C.}},
\bauthor{\bsnm{Zhang}, \binits{X.}},
\bauthor{\bsnm{Hu}, \binits{B.}}:
\batitle{A novel ensemble learning method using multiple objective particle
  swarm optimization for subject-independent eeg-based emotion recognition}.
\bjtitle{Computers in biology and medicine}
\bvolume{140},
\bfpage{105080}
(\byear{2022})
\end{barticle}
\endbibitem

\bibitem[\protect\citeauthoryear{Huang et~al.}{2021}]{huang2021differences}
\begin{barticle}
\bauthor{\bsnm{Huang}, \binits{D.}},
\bauthor{\bsnm{Chen}, \binits{S.}},
\bauthor{\bsnm{Liu}, \binits{C.}},
\bauthor{\bsnm{Zheng}, \binits{L.}},
\bauthor{\bsnm{Tian}, \binits{Z.}},
\bauthor{\bsnm{Jiang}, \binits{D.}}:
\batitle{Differences first in asymmetric brain: A bi-hemisphere discrepancy
  convolutional neural network for eeg emotion recognition}.
\bjtitle{Neurocomputing}
\bvolume{448},
\bfpage{140}--\blpage{151}
(\byear{2021})
\end{barticle}
\endbibitem

\bibitem[\protect\citeauthoryear{Algarni et~al.}{2022}]{algarni2022deep}
\begin{barticle}
\bauthor{\bsnm{Algarni}, \binits{M.}},
\bauthor{\bsnm{Saeed}, \binits{F.}},
\bauthor{\bsnm{Al-Hadhrami}, \binits{T.}},
\bauthor{\bsnm{Ghabban}, \binits{F.}},
\bauthor{\bsnm{Al-Sarem}, \binits{M.}}:
\batitle{Deep learning-based approach for emotion recognition using
  electroencephalography (eeg) signals using bi-directional long short-term
  memory (bi-lstm)}.
\bjtitle{Sensors}
\bvolume{22}(\bissue{8}),
\bfpage{2976}
(\byear{2022})
\end{barticle}
\endbibitem

\bibitem[\protect\citeauthoryear{Liao et~al.}{2024}]{liao2024exploring}
\begin{barticle}
\bauthor{\bsnm{Liao}, \binits{Y.}},
\bauthor{\bsnm{Gao}, \binits{Y.}},
\bauthor{\bsnm{Wang}, \binits{F.}},
\bauthor{\bsnm{Xu}, \binits{Z.}},
\bauthor{\bsnm{Wu}, \binits{Y.}},
\bauthor{\bsnm{Zhang}, \binits{L.}}:
\batitle{Exploring emotional experiences and dataset construction in the era of
  short videos based on physiological signals}.
\bjtitle{Biomedical Signal Processing and Control}
\bvolume{96},
\bfpage{106648}
(\byear{2024})
\end{barticle}
\endbibitem

\bibitem[\protect\citeauthoryear{Zhang et~al.}{2024}]{zhang2024tpro}
\begin{barticle}
\bauthor{\bsnm{Zhang}, \binits{X.}},
\bauthor{\bsnm{Cheng}, \binits{X.}},
\bauthor{\bsnm{Liu}, \binits{H.}}:
\batitle{Tpro-net: an eeg-based emotion recognition method reflecting subtle
  changes in emotion}.
\bjtitle{Scientific Reports}
\bvolume{14}(\bissue{1}),
\bfpage{13491}
(\byear{2024})
\end{barticle}
\endbibitem

\bibitem[\protect\citeauthoryear{Xu et~al.}{2024}]{xu2024infer}
\begin{barticle}
\bauthor{\bsnm{Xu}, \binits{C.}},
\bauthor{\bsnm{Liu}, \binits{L.}},
\bauthor{\bsnm{Jin}, \binits{L.}},
\bauthor{\bsnm{Du}, \binits{G.}},
\bauthor{\bsnm{Guo}, \binits{Z.}},
\bauthor{\bsnm{Zhao}, \binits{Y.}},
\bauthor{\bsnm{Huang}, \binits{X.}},
\bauthor{\bsnm{Li}, \binits{R.}}, \betal:
\batitle{Infer induced sentiment of comment response to video: A new task,
  dataset and baseline}.
\bjtitle{Advances in Neural Information Processing Systems}
\bvolume{37},
\bfpage{103737}--\blpage{103750}
(\byear{2024})
\end{barticle}
\endbibitem

\bibitem[\protect\citeauthoryear{Wagner et~al.}{2023}]{wagner2023dawn}
\begin{barticle}
\bauthor{\bsnm{Wagner}, \binits{J.}},
\bauthor{\bsnm{Triantafyllopoulos}, \binits{A.}},
\bauthor{\bsnm{Wierstorf}, \binits{H.}},
\bauthor{\bsnm{Schmitt}, \binits{M.}},
\bauthor{\bsnm{Burkhardt}, \binits{F.}},
\bauthor{\bsnm{Eyben}, \binits{F.}},
\bauthor{\bsnm{Schuller}, \binits{B.W.}}:
\batitle{Dawn of the transformer era in speech emotion recognition: closing the
  valence gap}.
\bjtitle{IEEE Transactions on Pattern Analysis and Machine Intelligence}
\bvolume{45}(\bissue{9}),
\bfpage{10745}--\blpage{10759}
(\byear{2023})
\end{barticle}
\endbibitem

\bibitem[\protect\citeauthoryear{Dudzik et~al.}{2020}]{dudzik2020blast}
\begin{botherref}
\oauthor{\bsnm{Dudzik}, \binits{B.}},
\oauthor{\bsnm{Broekens}, \binits{J.}},
\oauthor{\bsnm{Neerincx}, \binits{M.}},
\oauthor{\bsnm{Hung}, \binits{H.}}:
A blast from the past: Personalizing predictions of video-induced emotions
  using personal memories as context.
arXiv preprint arXiv:2008.12096
(2020)
\end{botherref}
\endbibitem

\bibitem[\protect\citeauthoryear{Yi et~al.}{2019}]{yi2019affective}
\begin{barticle}
\bauthor{\bsnm{Yi}, \binits{Y.}},
\bauthor{\bsnm{Wang}, \binits{H.}},
\bauthor{\bsnm{Li}, \binits{Q.}}:
\batitle{Affective video content analysis with adaptive fusion recurrent
  network}.
\bjtitle{IEEE Transactions on Multimedia}
\bvolume{22}(\bissue{9}),
\bfpage{2454}--\blpage{2466}
(\byear{2019})
\end{barticle}
\endbibitem

\bibitem[\protect\citeauthoryear{Xie et~al.}{2024}]{xie2024emovit}
\begin{bchapter}
\bauthor{\bsnm{Xie}, \binits{H.}},
\bauthor{\bsnm{Peng}, \binits{C.-J.}},
\bauthor{\bsnm{Tseng}, \binits{Y.-W.}},
\bauthor{\bsnm{Chen}, \binits{H.-J.}},
\bauthor{\bsnm{Hsu}, \binits{C.-F.}},
\bauthor{\bsnm{Shuai}, \binits{H.-H.}},
\bauthor{\bsnm{Cheng}, \binits{W.-H.}}:
\bctitle{Emovit: Revolutionizing emotion insights with visual instruction
  tuning}.
In: \bbtitle{Proceedings of the IEEE/CVF Conference on Computer Vision and
  Pattern Recognition},
pp. \bfpage{26596}--\blpage{26605}
(\byear{2024})
\end{bchapter}
\endbibitem

\bibitem[\protect\citeauthoryear{Li et~al.}{2024}]{li2024temporal}
\begin{bchapter}
\bauthor{\bsnm{Li}, \binits{X.}},
\bauthor{\bsnm{Wang}, \binits{S.}},
\bauthor{\bsnm{Huang}, \binits{X.}}:
\bctitle{Temporal enhancement for video affective content analysis}.
In: \bbtitle{Proceedings of the 32nd ACM International Conference on
  Multimedia},
pp. \bfpage{642}--\blpage{650}
(\byear{2024})
\end{bchapter}
\endbibitem

\bibitem[\protect\citeauthoryear{Chan and Jones}{2010}]{chan2010affect}
\begin{bchapter}
\bauthor{\bsnm{Chan}, \binits{C.H.}},
\bauthor{\bsnm{Jones}, \binits{G.J.}}:
\bctitle{An affect-based video retrieval system with open vocabulary querying}.
In: \bbtitle{International Workshop on Adaptive Multimedia Retrieval},
pp. \bfpage{103}--\blpage{117}
(\byear{2010}).
\bcomment{Springer}
\end{bchapter}
\endbibitem

\bibitem[\protect\citeauthoryear{Irie et~al.}{2010}]{irie2010affective}
\begin{barticle}
\bauthor{\bsnm{Irie}, \binits{G.}},
\bauthor{\bsnm{Satou}, \binits{T.}},
\bauthor{\bsnm{Kojima}, \binits{A.}},
\bauthor{\bsnm{Yamasaki}, \binits{T.}},
\bauthor{\bsnm{Aizawa}, \binits{K.}}:
\batitle{Affective audio-visual words and latent topic driving model for
  realizing movie affective scene classification}.
\bjtitle{IEEE Transactions on Multimedia}
\bvolume{12}(\bissue{6}),
\bfpage{523}--\blpage{535}
(\byear{2010})
\end{barticle}
\endbibitem

\bibitem[\protect\citeauthoryear{Kratzwald et~al.}{2018}]{kratzwald2018deep}
\begin{barticle}
\bauthor{\bsnm{Kratzwald}, \binits{B.}},
\bauthor{\bsnm{Ili{\'c}}, \binits{S.}},
\bauthor{\bsnm{Kraus}, \binits{M.}},
\bauthor{\bsnm{Feuerriegel}, \binits{S.}},
\bauthor{\bsnm{Prendinger}, \binits{H.}}:
\batitle{Deep learning for affective computing: Text-based emotion recognition
  in decision support}.
\bjtitle{Decision support systems}
\bvolume{115},
\bfpage{24}--\blpage{35}
(\byear{2018})
\end{barticle}
\endbibitem

\bibitem[\protect\citeauthoryear{Younis et~al.}{2024}]{younis2024machine}
\begin{barticle}
\bauthor{\bsnm{Younis}, \binits{E.M.}},
\bauthor{\bsnm{Mohsen}, \binits{S.}},
\bauthor{\bsnm{Houssein}, \binits{E.H.}},
\bauthor{\bsnm{Ibrahim}, \binits{O.A.S.}}:
\batitle{Machine learning for human emotion recognition: a comprehensive
  review}.
\bjtitle{Neural Computing and Applications}
\bvolume{36}(\bissue{16}),
\bfpage{8901}--\blpage{8947}
(\byear{2024})
\end{barticle}
\endbibitem

\bibitem[\protect\citeauthoryear{Hazmoune and
  Bougamouza}{2024}]{hazmoune2024using}
\begin{barticle}
\bauthor{\bsnm{Hazmoune}, \binits{S.}},
\bauthor{\bsnm{Bougamouza}, \binits{F.}}:
\batitle{Using transformers for multimodal emotion recognition: Taxonomies and
  state of the art review}.
\bjtitle{Engineering Applications of Artificial Intelligence}
\bvolume{133},
\bfpage{108339}
(\byear{2024})
\end{barticle}
\endbibitem

\bibitem[\protect\citeauthoryear{Zhang et~al.}{2024}]{zhang2024deep}
\begin{barticle}
\bauthor{\bsnm{Zhang}, \binits{S.}},
\bauthor{\bsnm{Yang}, \binits{Y.}},
\bauthor{\bsnm{Chen}, \binits{C.}},
\bauthor{\bsnm{Zhang}, \binits{X.}},
\bauthor{\bsnm{Leng}, \binits{Q.}},
\bauthor{\bsnm{Zhao}, \binits{X.}}:
\batitle{Deep learning-based multimodal emotion recognition from audio, visual,
  and text modalities: A systematic review of recent advancements and future
  prospects}.
\bjtitle{Expert Systems with Applications}
\bvolume{237},
\bfpage{121692}
(\byear{2024})
\end{barticle}
\endbibitem

\bibitem[\protect\citeauthoryear{Latif et~al.}{2021}]{latif2021survey}
\begin{barticle}
\bauthor{\bsnm{Latif}, \binits{S.}},
\bauthor{\bsnm{Rana}, \binits{R.}},
\bauthor{\bsnm{Khalifa}, \binits{S.}},
\bauthor{\bsnm{Jurdak}, \binits{R.}},
\bauthor{\bsnm{Qadir}, \binits{J.}},
\bauthor{\bsnm{Schuller}, \binits{B.}}:
\batitle{Survey of deep representation learning for speech emotion
  recognition}.
\bjtitle{IEEE Transactions on Affective Computing}
\bvolume{14}(\bissue{2}),
\bfpage{1634}--\blpage{1654}
(\byear{2021})
\end{barticle}
\endbibitem

\bibitem[\protect\citeauthoryear{Cui and Zheng}{2023}]{cui2023deep}
\begin{bbook}
\bauthor{\bsnm{Cui}, \binits{Z.}},
\bauthor{\bsnm{Zheng}, \binits{W.}}:
\bbtitle{Deep Learning Techniques Applied to Affective Computing}.
\bpublisher{Frontiers Media SA}, \blocation{???}
(\byear{2023})
\end{bbook}
\endbibitem

\bibitem[\protect\citeauthoryear{Kumar et~al.}{2024}]{kumar2024interpretable}
\begin{barticle}
\bauthor{\bsnm{Kumar}, \binits{P.}},
\bauthor{\bsnm{Malik}, \binits{S.}},
\bauthor{\bsnm{Raman}, \binits{B.}}:
\batitle{Interpretable multimodal emotion recognition using hybrid fusion of
  speech and image data}.
\bjtitle{Multimedia Tools and Applications}
\bvolume{83}(\bissue{10}),
\bfpage{28373}--\blpage{28394}
(\byear{2024})
\end{barticle}
\endbibitem

\bibitem[\protect\citeauthoryear{Corti{\~n}as-Lorenzo and
  Lacey}{2023}]{cortinas2023toward}
\begin{barticle}
\bauthor{\bsnm{Corti{\~n}as-Lorenzo}, \binits{K.}},
\bauthor{\bsnm{Lacey}, \binits{G.}}:
\batitle{Toward explainable affective computing: A review}.
\bjtitle{IEEE Transactions on Neural Networks and Learning Systems}
\bvolume{35}(\bissue{10}),
\bfpage{13101}--\blpage{13121}
(\byear{2023})
\end{barticle}
\endbibitem

\bibitem[\protect\citeauthoryear{Scherer et~al.}{2018}]{scherer2018appraisal}
\begin{barticle}
\bauthor{\bsnm{Scherer}, \binits{K.R.}},
\bauthor{\bsnm{Mortillaro}, \binits{M.}},
\bauthor{\bsnm{Rotondi}, \binits{I.}},
\bauthor{\bsnm{Sergi}, \binits{I.}},
\bauthor{\bsnm{Trznadel}, \binits{S.}}:
\batitle{Appraisal-driven facial actions as building blocks for emotion
  inference.}
\bjtitle{Journal of personality and social psychology}
\bvolume{114}(\bissue{3}),
\bfpage{358}
(\byear{2018})
\end{barticle}
\endbibitem

\bibitem[\protect\citeauthoryear{Lazarus}{1991}]{lazarus1991emotion}
\begin{bbook}
\bauthor{\bsnm{Lazarus}, \binits{R.S.}}:
\bbtitle{Emotion and Adaptation}.
\bpublisher{Oxford University Press},
\blocation{New York, NY}
(\byear{1991})
\end{bbook}
\endbibitem

\bibitem[\protect\citeauthoryear{Zall and
  Kangavari}{2022}]{zall2022comparative}
\begin{barticle}
\bauthor{\bsnm{Zall}, \binits{R.}},
\bauthor{\bsnm{Kangavari}, \binits{M.R.}}:
\batitle{Comparative analytical survey on cognitive agents with emotional
  intelligence}.
\bjtitle{Cognitive Computation}
\bvolume{14}(\bissue{4}),
\bfpage{1223}--\blpage{1246}
(\byear{2022})
\end{barticle}
\endbibitem

\bibitem[\protect\citeauthoryear{Zall et~al.}{2025}]{zall2025intelligent}
\begin{botherref}
\oauthor{\bsnm{Zall}, \binits{R.}},
\oauthor{\bsnm{Kheyrkhah}, \binits{A.}},
\oauthor{\bsnm{Cambria}, \binits{E.}},
\oauthor{\bsnm{Naseri}, \binits{Z.}},
\oauthor{\bsnm{Kangavari}, \binits{M.R.}}:
Intelligent agents with emotional intelligence: Current trends, challenges, and
  future prospects.
arXiv preprint arXiv:2511.20657
(2025)
\end{botherref}
\endbibitem

\bibitem[\protect\citeauthoryear{Zall and Kangavari}{2024}]{zall2024towards}
\begin{barticle}
\bauthor{\bsnm{Zall}, \binits{R.}},
\bauthor{\bsnm{Kangavari}, \binits{M.R.}}:
\batitle{Towards emotion-aware intelligent agents by utilizing knowledge graphs
  of experiences}.
\bjtitle{Cognitive Systems Research}
\bvolume{88},
\bfpage{101285}
(\byear{2024})
\end{barticle}
\endbibitem

\bibitem[\protect\citeauthoryear{Scherer et~al.}{2001}]{scherer2001appraisal}
\begin{bbook}
\bauthor{\bsnm{Scherer}, \binits{K.R.}},
\bauthor{\bsnm{Schorr}, \binits{A.}},
\bauthor{\bsnm{Johnstone}, \binits{T.}}:
\bbtitle{Appraisal Processes in Emotion: Theory, Methods, Research}.
\bpublisher{Oxford University Press}, \blocation{???}
(\byear{2001})
\end{bbook}
\endbibitem

\bibitem[\protect\citeauthoryear{Tian et~al.}{2017}]{tian2017recognizing}
\begin{bchapter}
\bauthor{\bsnm{Tian}, \binits{L.}},
\bauthor{\bsnm{Muszynski}, \binits{M.}},
\bauthor{\bsnm{Lai}, \binits{C.}},
\bauthor{\bsnm{Moore}, \binits{J.D.}},
\bauthor{\bsnm{Kostoulas}, \binits{T.}},
\bauthor{\bsnm{Lombardo}, \binits{P.}},
\bauthor{\bsnm{Pun}, \binits{T.}},
\bauthor{\bsnm{Chanel}, \binits{G.}}:
\bctitle{Recognizing induced emotions of movie audiences: Are induced and
  perceived emotions the same?}
In: \bbtitle{2017 Seventh International Conference on Affective Computing and
  Intelligent Interaction (ACII)},
pp. \bfpage{28}--\blpage{35}
(\byear{2017}).
\bcomment{IEEE}
\end{bchapter}
\endbibitem

\bibitem[\protect\citeauthoryear{Russell}{2003}]{russell2003core}
\begin{barticle}
\bauthor{\bsnm{Russell}, \binits{J.A.}}:
\batitle{Core affect and the psychological construction of emotion}.
\bjtitle{Psychological Review}
\bvolume{110}(\bissue{1}),
\bfpage{145}--\blpage{172}
(\byear{2003})
\doiurl{10.1037/0033-295X.110.1.145}
\end{barticle}
\endbibitem

\bibitem[\protect\citeauthoryear{Posner et~al.}{2005}]{posner2005circumplex}
\begin{barticle}
\bauthor{\bsnm{Posner}, \binits{J.}},
\bauthor{\bsnm{Russell}, \binits{J.A.}},
\bauthor{\bsnm{Peterson}, \binits{B.S.}}:
\batitle{The circumplex model of affect: An integrative approach to affective
  neuroscience, cognitive development, and psychopathology}.
\bjtitle{Development and psychopathology}
\bvolume{17}(\bissue{3}),
\bfpage{715}--\blpage{734}
(\byear{2005})
\end{barticle}
\endbibitem

\bibitem[\protect\citeauthoryear{Mehrabian}{1996}]{mehrabian1996pleasure}
\begin{barticle}
\bauthor{\bsnm{Mehrabian}, \binits{A.}}:
\batitle{Pleasure-arousal-dominance: A general framework for describing and
  measuring individual differences in temperament}.
\bjtitle{Current psychology}
\bvolume{14}(\bissue{4}),
\bfpage{261}--\blpage{292}
(\byear{1996})
\end{barticle}
\endbibitem

\bibitem[\protect\citeauthoryear{Bakker et~al.}{2014}]{bakker2014pleasure}
\begin{barticle}
\bauthor{\bsnm{Bakker}, \binits{I.}},
\bauthor{\bsnm{Van Der~Voordt}, \binits{T.}},
\bauthor{\bsnm{Vink}, \binits{P.}},
\bauthor{\bsnm{De~Boon}, \binits{J.}}:
\batitle{Pleasure, arousal, dominance: Mehrabian and russell revisited}.
\bjtitle{Current psychology}
\bvolume{33}(\bissue{3}),
\bfpage{405}--\blpage{421}
(\byear{2014})
\end{barticle}
\endbibitem

\bibitem[\protect\citeauthoryear{Wrobel}{2025}]{wrobel2025proxy}
\begin{botherref}
\oauthor{\bsnm{Wrobel}, \binits{M.R.}}:
A proxy-based method for mapping discrete emotions onto vad model.
arXiv preprint arXiv:2511.12521
(2025)
\end{botherref}
\endbibitem

\bibitem[\protect\citeauthoryear{Jia et~al.}{2025}]{jia2025bridging}
\begin{bchapter}
\bauthor{\bsnm{Jia}, \binits{J.}},
\bauthor{\bsnm{Zhang}, \binits{H.}},
\bauthor{\bsnm{Liang}, \binits{J.}}:
\bctitle{Bridging discrete and continuous: A multimodal strategy for complex
  emotion detection}.
In: \bbtitle{2025 IEEE 35th International Workshop on Machine Learning for
  Signal Processing (MLSP)},
pp. \bfpage{1}--\blpage{6}
(\byear{2025}).
\bcomment{IEEE}
\end{bchapter}
\endbibitem

\bibitem[\protect\citeauthoryear{Somarathna
  et~al.}{2023}]{somarathna2023emostim}
\begin{barticle}
\bauthor{\bsnm{Somarathna}, \binits{R.}},
\bauthor{\bsnm{Vuilleumier}, \binits{P.}},
\bauthor{\bsnm{Mohammadi}, \binits{G.}}:
\batitle{Emostim: A database of emotional film clips with discrete and
  componential assessment}.
\bjtitle{IEEE Transactions on Affective Computing}
\bvolume{15}(\bissue{3}),
\bfpage{1202}--\blpage{1212}
(\byear{2023})
\end{barticle}
\endbibitem

\bibitem[\protect\citeauthoryear{Moors et~al.}{2013}]{moors2013appraisal}
\begin{barticle}
\bauthor{\bsnm{Moors}, \binits{A.}},
\bauthor{\bsnm{Ellsworth}, \binits{P.C.}},
\bauthor{\bsnm{Scherer}, \binits{K.R.}},
\bauthor{\bsnm{Frijda}, \binits{N.H.}}:
\batitle{Appraisal theories of emotion: State of the art and future
  development}.
\bjtitle{Emotion review}
\bvolume{5}(\bissue{2}),
\bfpage{119}--\blpage{124}
(\byear{2013})
\end{barticle}
\endbibitem

\bibitem[\protect\citeauthoryear{Sullins
  et~al.}{2024}]{sullins2024investigating}
\begin{barticle}
\bauthor{\bsnm{Sullins}, \binits{J.}},
\bauthor{\bsnm{Turner}, \binits{J.}},
\bauthor{\bsnm{Kim}, \binits{J.}},
\bauthor{\bsnm{Barber}, \binits{S.}}:
\batitle{Investigating the impacts of shame-proneness on students’ state
  shame, self-regulation, and learning}.
\bjtitle{Education Sciences}
\bvolume{14}(\bissue{2}),
\bfpage{138}
(\byear{2024})
\end{barticle}
\endbibitem

\bibitem[\protect\citeauthoryear{Soleymani}{2016}]{soleymani2016detecting}
\begin{botherref}
\oauthor{\bsnm{Soleymani}, \binits{M.}}:
Detecting cognitive appraisals from facial expressions for interest
  recognition.
arXiv preprint arXiv:1609.09761
(2016)
\end{botherref}
\endbibitem

\bibitem[\protect\citeauthoryear{Barradas et~al.}{2025}]{barradas2025dynamic}
\begin{barticle}
\bauthor{\bsnm{Barradas}, \binits{I.}},
\bauthor{\bsnm{Tschiesner}, \binits{R.}},
\bauthor{\bsnm{Peer}, \binits{A.}}:
\batitle{Dynamic emotion intensity estimation from physiological signals
  facilitating interpretation via appraisal theory}.
\bjtitle{PloS one}
\bvolume{20}(\bissue{1}),
\bfpage{0315929}
(\byear{2025})
\end{barticle}
\endbibitem

\bibitem[\protect\citeauthoryear{Frijda et~al.}{1989}]{frijda1989relations}
\begin{barticle}
\bauthor{\bsnm{Frijda}, \binits{N.H.}},
\bauthor{\bsnm{Kuipers}, \binits{P.}},
\bauthor{\bsnm{Ter~Schure}, \binits{E.}}:
\batitle{Relations among emotion, appraisal, and emotional action readiness.}
\bjtitle{Journal of personality and social psychology}
\bvolume{57}(\bissue{2}),
\bfpage{212}
(\byear{1989})
\end{barticle}
\endbibitem

\bibitem[\protect\citeauthoryear{Scherer}{1993}]{scherer1993studying}
\begin{barticle}
\bauthor{\bsnm{Scherer}, \binits{K.R.}}:
\batitle{Studying the emotion-antecedent appraisal process: An expert system
  approach}.
\bjtitle{Cognition \& Emotion}
\bvolume{7}(\bissue{3-4}),
\bfpage{325}--\blpage{355}
(\byear{1993})
\end{barticle}
\endbibitem

\bibitem[\protect\citeauthoryear{Ortony et~al.}{2022}]{ortony2022cognitive}
\begin{bbook}
\bauthor{\bsnm{Ortony}, \binits{A.}},
\bauthor{\bsnm{Clore}, \binits{G.L.}},
\bauthor{\bsnm{Collins}, \binits{A.}}:
\bbtitle{The Cognitive Structure of Emotions}.
\bpublisher{Cambridge university press}, \blocation{???}
(\byear{2022})
\end{bbook}
\endbibitem

\bibitem[\protect\citeauthoryear{Juslin}{2025}]{juslin2025major}
\begin{barticle}
\bauthor{\bsnm{Juslin}, \binits{P.N.}}:
\batitle{Major theories of emotion causation and their applicability to music:
  The case for multi-level approaches}.
\bjtitle{Music Perception: An Interdisciplinary Journal}
\bvolume{42}(\bissue{5}),
\bfpage{421}--\blpage{466}
(\byear{2025})
\end{barticle}
\endbibitem

\bibitem[\protect\citeauthoryear{Friedrich
  et~al.}{2025}]{friedrich2025emotional}
\begin{botherref}
\oauthor{\bsnm{Friedrich}, \binits{T.L.}},
\oauthor{\bsnm{Kiefer}, \binits{T.}},
\oauthor{\bsnm{Eubanks}, \binits{D.}}:
Emotional reactions to idea evaluation: Creative perseverance and evaluating
  others.
Creativity Research Journal,
1--21
(2025)
\end{botherref}
\endbibitem

\bibitem[\protect\citeauthoryear{Castellanos
  et~al.}{2026}]{castellanos2026systematic}
\begin{botherref}
\oauthor{\bsnm{Castellanos}, \binits{S.}},
\oauthor{\bsnm{Cuen}, \binits{E.O.}},
\oauthor{\bsnm{Padilla}, \binits{E.L.}},
\oauthor{\bsnm{Rodr{\'\i}guez}, \binits{L.-F.}}:
Systematic guidelines for extending the appraisal process in computational
  models of emotion.
Cognitive Systems Research,
101442
(2026)
\end{botherref}
\endbibitem

\bibitem[\protect\citeauthoryear{Tak et~al.}{2025}]{tak2025aware}
\begin{botherref}
\oauthor{\bsnm{Tak}, \binits{A.N.}},
\oauthor{\bsnm{Gratch}, \binits{J.}},
\oauthor{\bsnm{Scherer}, \binits{K.R.}}:
Aware yet biased: Investigating emotional reasoning and appraisal bias in large
  language models.
IEEE Transactions on Affective Computing
(2025)
\end{botherref}
\endbibitem

\bibitem[\protect\citeauthoryear{Xu et~al.}{2025}]{xu2025conversational}
\begin{botherref}
\oauthor{\bsnm{Xu}, \binits{K.}},
\oauthor{\bsnm{Xie}, \binits{C.}},
\oauthor{\bsnm{Liu}, \binits{Q.}},
\oauthor{\bsnm{Du}, \binits{Y.}},
\oauthor{\bsnm{Li}, \binits{X.}},
\oauthor{\bsnm{Li}, \binits{Y.}},
\oauthor{\bsnm{Liu}, \binits{J.}}:
Conversational emotion prediction based on appraisal theory: K. xu et al.
Soft Computing,
1--15
(2025)
\end{botherref}
\endbibitem

\bibitem[\protect\citeauthoryear{Krzeminska}{2025}]{krzeminska2025}
\begin{barticle}
\bauthor{\bsnm{Krzeminska}, \binits{I.}}:
\batitle{Multimodal recognition of users states at human-ai interaction
  adaptation}.
\bjtitle{Technium Romanian Journal of Applied Sciences and Technology}
\bvolume{26},
\bfpage{102}--\blpage{140}
(\byear{2025})
\doiurl{10.47577/technium.v26i.12398}
\end{barticle}
\endbibitem

\bibitem[\protect\citeauthoryear{Ekman}{1992}]{ekman1992argument}
\begin{barticle}
\bauthor{\bsnm{Ekman}, \binits{P.}}:
\batitle{An argument for basic emotions}.
\bjtitle{Cognition \& emotion}
\bvolume{6}(\bissue{3-4}),
\bfpage{169}--\blpage{200}
(\byear{1992})
\end{barticle}
\endbibitem

\bibitem[\protect\citeauthoryear{Russell}{1980}]{russell1980circumplex}
\begin{barticle}
\bauthor{\bsnm{Russell}, \binits{J.A.}}:
\batitle{A circumplex model of affect.}
\bjtitle{Journal of personality and social psychology}
\bvolume{39}(\bissue{6}),
\bfpage{1161}
(\byear{1980})
\end{barticle}
\endbibitem

\bibitem[\protect\citeauthoryear{Reisenzein}{1994}]{reisenzein1994pleasure}
\begin{barticle}
\bauthor{\bsnm{Reisenzein}, \binits{R.}}:
\batitle{Pleasure-arousal theory and the intensity of emotions.}
\bjtitle{Journal of personality and social psychology}
\bvolume{67}(\bissue{3}),
\bfpage{525}
(\byear{1994})
\end{barticle}
\endbibitem

\bibitem[\protect\citeauthoryear{Taverner et~al.}{2021}]{taverner2021fuzzy}
\begin{barticle}
\bauthor{\bsnm{Taverner}, \binits{J.}},
\bauthor{\bsnm{Vivancos}, \binits{E.}},
\bauthor{\bsnm{Botti}, \binits{V.}}:
\batitle{A fuzzy appraisal model for affective agents adapted to cultural
  environments using the pleasure and arousal dimensions}.
\bjtitle{Information Sciences}
\bvolume{546},
\bfpage{74}--\blpage{86}
(\byear{2021})
\end{barticle}
\endbibitem

\bibitem[\protect\citeauthoryear{Li et~al.}{2020}]{li2020exploring}
\begin{barticle}
\bauthor{\bsnm{Li}, \binits{C.}},
\bauthor{\bsnm{Bao}, \binits{Z.}},
\bauthor{\bsnm{Li}, \binits{L.}},
\bauthor{\bsnm{Zhao}, \binits{Z.}}:
\batitle{Exploring temporal representations by leveraging attention-based
  bidirectional lstm-rnns for multi-modal emotion recognition}.
\bjtitle{Information Processing \& Management}
\bvolume{57}(\bissue{3}),
\bfpage{102185}
(\byear{2020})
\end{barticle}
\endbibitem

\bibitem[\protect\citeauthoryear{Ahire et~al.}{2025}]{ahire2025maven}
\begin{bchapter}
\bauthor{\bsnm{Ahire}, \binits{V.}},
\bauthor{\bsnm{Shah}, \binits{K.}},
\bauthor{\bsnm{Khan}, \binits{M.}},
\bauthor{\bsnm{Pakhale}, \binits{N.}},
\bauthor{\bsnm{Sookha}, \binits{L.}},
\bauthor{\bsnm{Ganaie}, \binits{M.}},
\bauthor{\bsnm{Dhall}, \binits{A.}}:
\bctitle{Maven: Multi-modal attention for valence-arousal emotion network}.
In: \bbtitle{Proceedings of the Computer Vision and Pattern Recognition
  Conference},
pp. \bfpage{5789}--\blpage{5799}
(\byear{2025})
\end{bchapter}
\endbibitem

\bibitem[\protect\citeauthoryear{Hu et~al.}{2026}]{hu2026target}
\begin{botherref}
\oauthor{\bsnm{Hu}, \binits{Q.}},
\oauthor{\bsnm{Murad}, \binits{M.A.A.}},
\oauthor{\bsnm{Azman}, \binits{A.B.}},
\oauthor{\bsnm{Nasharuddin}, \binits{N.A.}}:
Target-conditioned triple-path consistency for distributional music emotion
  regression.
Knowledge-Based Systems,
115317
(2026)
\end{botherref}
\endbibitem

\bibitem[\protect\citeauthoryear{Liu and Yang}{2025}]{liu2025emotional}
\begin{bchapter}
\bauthor{\bsnm{Liu}, \binits{Y.}},
\bauthor{\bsnm{Yang}, \binits{J.}}:
\bctitle{Emotional speech synthesis based on valence-arousal-dominance model
  and multi-feature codebook}.
In: \bbtitle{2025 International Conference on Asian Language Processing
  (IALP)},
pp. \bfpage{255}--\blpage{259}
(\byear{2025}).
\bcomment{IEEE}
\end{bchapter}
\endbibitem

\bibitem[\protect\citeauthoryear{Yu et~al.}{2026}]{yu2026emotion}
\begin{barticle}
\bauthor{\bsnm{Yu}, \binits{Y.}},
\bauthor{\bsnm{Xu}, \binits{H.}},
\bauthor{\bsnm{Xu}, \binits{Z.}},
\bauthor{\bsnm{Duan}, \binits{Y.}},
\bauthor{\bsnm{Wang}, \binits{R.}},
\bauthor{\bsnm{Zheng}, \binits{H.}},
\bauthor{\bsnm{Li}, \binits{Y.}},
\bauthor{\bsnm{Xu}, \binits{Y.}}:
\batitle{An emotion recognition approach using peripheral physiological signals
  based on hierarchical gated residuals and receptive field attention}.
\bjtitle{Engineering Applications of Artificial Intelligence}
\bvolume{166},
\bfpage{113346}
(\byear{2026})
\end{barticle}
\endbibitem

\bibitem[\protect\citeauthoryear{Priyadarshani and
  Miyapuram}{2024}]{priyadarshani2024predicting}
\begin{bchapter}
\bauthor{\bsnm{Priyadarshani}, \binits{M.}},
\bauthor{\bsnm{Miyapuram}, \binits{K.P.}}:
\bctitle{Predicting valence and arousal from affective images: A comparative
  analysis of deep learning and random forest regressor}.
In: \bbtitle{Proceedings of the 8th International Conference on Data Science
  and Management of Data (12th ACM IKDD CODS and 30th COMAD)},
pp. \bfpage{350}--\blpage{352}
(\byear{2024})
\end{bchapter}
\endbibitem

\bibitem[\protect\citeauthoryear{Thao et~al.}{2021}]{thao2021attendaffectnet}
\begin{barticle}
\bauthor{\bsnm{Thao}, \binits{H.T.P.}},
\bauthor{\bsnm{Balamurali}, \binits{B.}},
\bauthor{\bsnm{Roig}, \binits{G.}},
\bauthor{\bsnm{Herremans}, \binits{D.}}:
\batitle{Attendaffectnet--emotion prediction of movie viewers using multimodal
  fusion with self-attention}.
\bjtitle{Sensors}
\bvolume{21}(\bissue{24}),
\bfpage{8356}
(\byear{2021})
\end{barticle}
\endbibitem

\bibitem[\protect\citeauthoryear{Zhang et~al.}{2024}]{zhang2024mart}
\begin{bchapter}
\bauthor{\bsnm{Zhang}, \binits{Z.}},
\bauthor{\bsnm{Zhao}, \binits{P.}},
\bauthor{\bsnm{Park}, \binits{E.}},
\bauthor{\bsnm{Yang}, \binits{J.}}:
\bctitle{Mart: Masked affective representation learning via masked temporal
  distribution distillation}.
In: \bbtitle{Proceedings of the IEEE/CVF Conference on Computer Vision and
  Pattern Recognition},
pp. \bfpage{12830}--\blpage{12840}
(\byear{2024})
\end{bchapter}
\endbibitem

\bibitem[\protect\citeauthoryear{Guo et~al.}{2025}]{guo2025stimuvar}
\begin{botherref}
\oauthor{\bsnm{Guo}, \binits{Y.}},
\oauthor{\bsnm{Siddiqui}, \binits{F.}},
\oauthor{\bsnm{Zhao}, \binits{Y.}},
\oauthor{\bsnm{Chellappa}, \binits{R.}},
\oauthor{\bsnm{Lo}, \binits{S.-Y.}}:
Stimuvar: Spatiotemporal stimuli-aware video affective reasoning with
  multimodal large language models.
International Journal of Computer Vision,
1--17
(2025)
\end{botherref}
\endbibitem

\bibitem[\protect\citeauthoryear{Zhang et~al.}{2025}]{zhang2025videmo}
\begin{botherref}
\oauthor{\bsnm{Zhang}, \binits{Z.}},
\oauthor{\bsnm{Wang}, \binits{W.}},
\oauthor{\bsnm{Zhu}, \binits{Y.}},
\oauthor{\bsnm{Qin}, \binits{W.}},
\oauthor{\bsnm{Wan}, \binits{P.}},
\oauthor{\bsnm{Zhang}, \binits{D.}},
\oauthor{\bsnm{Yang}, \binits{J.}}:
Videmo: Affective-tree reasoning for emotion-centric video foundation models.
arXiv preprint arXiv:2511.02712
(2025)
\end{botherref}
\endbibitem

\bibitem[\protect\citeauthoryear{Lian et~al.}{2025}]{lian2025mer}
\begin{bchapter}
\bauthor{\bsnm{Lian}, \binits{Z.}},
\bauthor{\bsnm{Liu}, \binits{R.}},
\bauthor{\bsnm{Xu}, \binits{K.}},
\bauthor{\bsnm{Liu}, \binits{B.}},
\bauthor{\bsnm{Liu}, \binits{X.}},
\bauthor{\bsnm{Zhang}, \binits{Y.}},
\bauthor{\bsnm{Liu}, \binits{X.}},
\bauthor{\bsnm{Li}, \binits{Y.}},
\bauthor{\bsnm{Cheng}, \binits{Z.}},
\bauthor{\bsnm{Zuo}, \binits{H.}}, \betal:
\bctitle{Mer 2025: When affective computing meets large language models}.
In: \bbtitle{Proceedings of the 33rd ACM International Conference on
  Multimedia},
pp. \bfpage{13837}--\blpage{13842}
(\byear{2025})
\end{bchapter}
\endbibitem

\bibitem[\protect\citeauthoryear{Topic and Russo}{2021}]{topic2021emotion}
\begin{barticle}
\bauthor{\bsnm{Topic}, \binits{A.}},
\bauthor{\bsnm{Russo}, \binits{M.}}:
\batitle{Emotion recognition based on eeg feature maps through deep learning
  network}.
\bjtitle{Engineering Science and Technology, an International Journal}
\bvolume{24}(\bissue{6}),
\bfpage{1442}--\blpage{1454}
(\byear{2021})
\end{barticle}
\endbibitem

\bibitem[\protect\citeauthoryear{Dudzik et~al.}{2020}]{dudzik2020exploring}
\begin{bchapter}
\bauthor{\bsnm{Dudzik}, \binits{B.}},
\bauthor{\bsnm{Broekens}, \binits{J.}},
\bauthor{\bsnm{Neerincx}, \binits{M.}},
\bauthor{\bsnm{Hung}, \binits{H.}}:
\bctitle{Exploring personal memories and video content as context for facial
  behavior in predictions of video-induced emotions}.
In: \bbtitle{Proceedings of the 2020 International Conference on Multimodal
  Interaction},
pp. \bfpage{153}--\blpage{162}
(\byear{2020})
\end{bchapter}
\endbibitem

\bibitem[\protect\citeauthoryear{Antonov et~al.}{2024}]{antonov2024decoding}
\begin{barticle}
\bauthor{\bsnm{Antonov}, \binits{A.}},
\bauthor{\bsnm{Kumar}, \binits{S.S.}},
\bauthor{\bsnm{Wei}, \binits{J.}},
\bauthor{\bsnm{Headley}, \binits{W.}},
\bauthor{\bsnm{Wood}, \binits{O.}},
\bauthor{\bsnm{Montana}, \binits{G.}}:
\batitle{Decoding viewer emotions in video ads}.
\bjtitle{Scientific Reports}
\bvolume{14}(\bissue{1}),
\bfpage{26382}
(\byear{2024})
\end{barticle}
\endbibitem

\bibitem[\protect\citeauthoryear{Dudzik et~al.}{2021}]{dudzik2021collecting}
\begin{barticle}
\bauthor{\bsnm{Dudzik}, \binits{B.}},
\bauthor{\bsnm{Hung}, \binits{H.}},
\bauthor{\bsnm{Neerincx}, \binits{M.}},
\bauthor{\bsnm{Broekens}, \binits{J.}}:
\batitle{Collecting mementos: A multimodal dataset for context-sensitive
  modeling of affect and memory processing in responses to videos}.
\bjtitle{IEEE Transactions on Affective Computing}
\bvolume{14}(\bissue{2}),
\bfpage{1249}--\blpage{1266}
(\byear{2021})
\end{barticle}
\endbibitem

\bibitem[\protect\citeauthoryear{Kamran et~al.}{2023}]{kamran2023emodnn}
\begin{barticle}
\bauthor{\bsnm{Kamran}, \binits{S.}},
\bauthor{\bsnm{Zall}, \binits{R.}},
\bauthor{\bsnm{Hosseini}, \binits{S.}},
\bauthor{\bsnm{Kangavari}, \binits{M.}},
\bauthor{\bsnm{Rahmani}, \binits{S.}},
\bauthor{\bsnm{Hua}, \binits{W.}}:
\batitle{Emodnn: understanding emotions from short texts through a deep neural
  network ensemble}.
\bjtitle{Neural Computing and Applications}
\bvolume{35}(\bissue{18}),
\bfpage{13565}--\blpage{13582}
(\byear{2023})
\end{barticle}
\endbibitem

\bibitem[\protect\citeauthoryear{Rahmani et~al.}{2023}]{rahmani2023transfer}
\begin{barticle}
\bauthor{\bsnm{Rahmani}, \binits{S.}},
\bauthor{\bsnm{Hosseini}, \binits{S.}},
\bauthor{\bsnm{Zall}, \binits{R.}},
\bauthor{\bsnm{Kangavari}, \binits{M.R.}},
\bauthor{\bsnm{Kamran}, \binits{S.}},
\bauthor{\bsnm{Hua}, \binits{W.}}:
\batitle{Transfer-based adaptive tree for multimodal sentiment analysis based
  on user latent aspects}.
\bjtitle{Knowledge-Based Systems}
\bvolume{261},
\bfpage{110219}
(\byear{2023})
\end{barticle}
\endbibitem

\bibitem[\protect\citeauthoryear{Sander et~al.}{2005}]{sander2005systems}
\begin{barticle}
\bauthor{\bsnm{Sander}, \binits{D.}},
\bauthor{\bsnm{Grandjean}, \binits{D.}},
\bauthor{\bsnm{Scherer}, \binits{K.R.}}:
\batitle{A systems approach to appraisal mechanisms in emotion}.
\bjtitle{Neural networks}
\bvolume{18}(\bissue{4}),
\bfpage{317}--\blpage{352}
(\byear{2005})
\end{barticle}
\endbibitem

\bibitem[\protect\citeauthoryear{Vaswani et~al.}{2017}]{vaswani2017attention}
\begin{botherref}
\oauthor{\bsnm{Vaswani}, \binits{A.}},
\oauthor{\bsnm{Shazeer}, \binits{N.}},
\oauthor{\bsnm{Parmar}, \binits{N.}},
\oauthor{\bsnm{Uszkoreit}, \binits{J.}},
\oauthor{\bsnm{Jones}, \binits{L.}},
\oauthor{\bsnm{Gomez}, \binits{A.N.}},
\oauthor{\bsnm{Kaiser}, \binits{{\L}.}},
\oauthor{\bsnm{Polosukhin}, \binits{I.}}:
Attention is all you need.
Advances in neural information processing systems
\textbf{30}
(2017)
\end{botherref}
\endbibitem

\bibitem[\protect\citeauthoryear{Taverner et~al.}{2019}]{taverner2019towards}
\begin{bchapter}
\bauthor{\bsnm{Taverner}, \binits{J.}},
\bauthor{\bsnm{Vivancos}, \binits{E.}},
\bauthor{\bsnm{Botti}, \binits{V.J.}}:
\bctitle{Towards a computational approach to emotion elicitation in affective
  agents.}
In: \bbtitle{ICAART (1)},
pp. \bfpage{275}--\blpage{280}
(\byear{2019})
\end{bchapter}
\endbibitem

\bibitem[\protect\citeauthoryear{Baevski et~al.}{2020}]{baevski2020wav2vec}
\begin{barticle}
\bauthor{\bsnm{Baevski}, \binits{A.}},
\bauthor{\bsnm{Zhou}, \binits{Y.}},
\bauthor{\bsnm{Mohamed}, \binits{A.}},
\bauthor{\bsnm{Auli}, \binits{M.}}:
\batitle{wav2vec 2.0: A framework for self-supervised learning of speech
  representations}.
\bjtitle{Advances in neural information processing systems}
\bvolume{33},
\bfpage{12449}--\blpage{12460}
(\byear{2020})
\end{barticle}
\endbibitem

\bibitem[\protect\citeauthoryear{Baevski et~al.}{2022}]{baevski2022data2vec}
\begin{bchapter}
\bauthor{\bsnm{Baevski}, \binits{A.}},
\bauthor{\bsnm{Hsu}, \binits{W.-N.}},
\bauthor{\bsnm{Xu}, \binits{Q.}},
\bauthor{\bsnm{Babu}, \binits{A.}},
\bauthor{\bsnm{Gu}, \binits{J.}},
\bauthor{\bsnm{Auli}, \binits{M.}}:
\bctitle{Data2vec: A general framework for self-supervised learning in speech,
  vision and language}.
In: \bbtitle{International Conference on Machine Learning},
pp. \bfpage{1298}--\blpage{1312}
(\byear{2022}).
\bcomment{PMLR}
\end{bchapter}
\endbibitem

\bibitem[\protect\citeauthoryear{Bertasius et~al.}{2021}]{bertasius2021space}
\begin{bchapter}
\bauthor{\bsnm{Bertasius}, \binits{G.}},
\bauthor{\bsnm{Wang}, \binits{H.}},
\bauthor{\bsnm{Torresani}, \binits{L.}}:
\bctitle{Is space-time attention all you need for video understanding?}
In: \bbtitle{Icml},
vol. \bseriesno{2},
p. \bfpage{4}
(\byear{2021})
\end{bchapter}
\endbibitem

\bibitem[\protect\citeauthoryear{Dosovitskiy}{2020}]{dosovitskiy2020image}
\begin{botherref}
\oauthor{\bsnm{Dosovitskiy}, \binits{A.}}:
An image is worth 16x16 words: Transformers for image recognition at scale.
arXiv preprint arXiv:2010.11929
(2020)
\end{botherref}
\endbibitem

\bibitem[\protect\citeauthoryear{Loshchilov and
  Hutter}{2017}]{loshchilov2017decoupled}
\begin{botherref}
\oauthor{\bsnm{Loshchilov}, \binits{I.}},
\oauthor{\bsnm{Hutter}, \binits{F.}}:
Decoupled weight decay regularization.
arXiv preprint arXiv:1711.05101
(2017)
\end{botherref}
\endbibitem

\end{thebibliography}

\end{document}